\documentclass{article}

\usepackage[english]{babel}

\usepackage[letterpaper,top=2cm,bottom=2cm,left=3cm,right=3cm,marginparwidth=1.75cm]{geometry}

\usepackage{amsmath}
\usepackage{graphicx}
\usepackage[colorlinks=true, allcolors=blue]{hyperref}
\usepackage{authblk}
\usepackage[T1]{fontenc}

\title{Large Language Models for Manufacturing}
\author[1$^*$]{Yiwei Li}
\author[1$^*$]{Huaqin Zhao}
\author[1$^*$]{Hanqi Jiang}
\author[1$^*$]{Yi Pan}
\author[1$^*$]{Zhengliang Liu}
\author[1$^*$]{Zihao Wu}
\author[1$^*$]{Peng Shu}
\author[2$^*$]{Jie Tian}
\author[1$^*$]{Tianze Yang}
\author[1$^*$]{Shaochen Xu}
\author[3$^*$]{Yanjun Lyu}
\author[2]{Parker Blenk}
\author[2]{Jacob Pence}
\author[2]{Jason Rupram}
\author[2]{Eliza Banu}
\author[1]{Ninghao Liu}
\author[2]{Linbing Wang}
\author[4]{Wenzhan Song}
\author[5]{Xiaoming Zhai}
\author[2]{Kenan Song}
\author[3]{Dajiang Zhu}
\author[2]{Beiwen Li}
\author[2]{Xianqiao Wang}
\author[1$\dagger$]{Tianming Liu}

\affil[1]{School of Computing, University of Georgia, USA}
\affil[2]{School of Environmental, Civil, Agricultural and Mechanical Engineering, University of Georgia, USA}
\affil[3]{Department of Computer Science and Engineering, University of Texas at Arlington, USA}
\affil[4]{School of Electrical and Computer Engineering, University of Georgia, USA}
\affil[5]{Department of Mathematics, Science, and Social Studies Education, University of Georgia, USA}

\begin{document}
\maketitle
\thanks{$^*$Equal contribution.}
\thanks{$\dagger$Corresponding authors.}


\begin{abstract}
The rapid advances in Large Language Models (LLMs) have the potential to transform manufacturing industry, offering new opportunities to optimize processes, improve efficiency, and drive innovation. This paper provides a comprehensive exploration of the integration of LLMs into the manufacturing domain, focusing on their potential to automate and enhance various aspects of manufacturing, from product design and development to quality control, supply chain optimization, and talent management. Through extensive evaluations across multiple manufacturing tasks, we demonstrate the remarkable capabilities of state-of-the-art LLMs, such as GPT-4V, in understanding and executing complex instructions, extracting valuable insights from vast amounts of data, and facilitating knowledge sharing. We also delve into the transformative potential of LLMs in reshaping manufacturing education, automating coding processes, enhancing robot control systems, and enabling the creation of immersive, data-rich virtual environments through the industrial metaverse. By highlighting the practical applications and emerging use cases of LLMs in manufacturing, this paper aims to provide a valuable resource for professionals, researchers, and decision-makers seeking to harness the power of these technologies to address real-world challenges, drive operational excellence, and unlock sustainable growth in an increasingly competitive landscape.
\end{abstract}

\section{Introduction}
In recent years, the advent of Large Language Models (LLMs) such as OpenAI's GPT family~\cite{openaiIntroducingChatGPT} has heralded significant advancements in the domain of Natural Language Processing (NLP). These developments represent a pivotal milestone in AI technology, particularly in its capacity to comprehend and generate natural language. Bolstered by enhanced computational resources and refined algorithms, LLMs have showcased remarkable proficiency in contextual comprehension, question answering, and content generation. Notably, within the domain of manufacturing, these capabilities are progressively unveiling their immense potential~\cite{chang2023survey,wu2023brief,wu2023bloomberggpt}.

Manufacturing, as an extremely technical field, requires various technologies to improve efficiency, quality, and innovation capabilities~\cite{bikas2016additive}. For example, advanced manufacturing technologies including 3D printing, robotics, automated production lines, etc.~\cite{karayel2020additive} are needed to improve production efficiency, flexibility, and precision~\cite{pereira2019comparison}. Digital manufacturing uses the industrial Internet, big data analysis, AI, and other technologies to achieve diverse, intelligent, and visual management of the production process~\cite{wang2018deep}. Virtual simulation technology optimizes and verifies product design and production processes through computer simulation and virtual reality technology~\cite{han2020recent}. There are also manufacturing sectors that require a lot of text-processing work in the fields of project application, project review, program research, patent management, etc. 

Now the implementation of these technologies can bring about a qualitative leap through the application of LLM technology. LLM has the potential to bring huge changes to the entire industry because of its powerful logical reasoning, knowledge transfer, and text-processing capabilities. 

The deployment of LLMs within the manufacturing sector encounters several substantial challenges~\cite{liu2024surviving,sun2024trustllm,liu2024understanding,zhao2023ophtha,holmes2023evaluating}. Firstly, the manufacturing domain encompasses highly specialized and intricate data~\cite{liu2024radiation}, characterized by specific terminologies, regulatory frameworks, and dynamic market conditions~\cite{liao2023differentiating}. These aspects demand sophisticated comprehension capabilities from the models. Given the proximity of manufacturing to engineering disciplines, the expectations for outcomes generated by LLMs in this field are notably high~\cite{lee2023multimodality}. While it is widely acknowledged that LLMs possess significant potential and capabilities for logical reasoning~\cite{liu2023tailoring}, the efficient and effective application of these models to various downstream tasks in manufacturing presents a substantial challenge that merits careful consideration.

To mitigate these challenges, ongoing efforts are directed towards enhancing the algorithms underpinning LLMs to refine their proficiency in processing and understanding domain-specific knowledge~\cite{liu2024radiation,liu2023tailoring}. By training these models on a substantial corpus of domain-specific data, they can better assimilate specialized knowledge pertinent to manufacturing and make them more suitable for related applications. Moreover, integrating expert systems and implementing manual review protocols can further bolster the accuracy and dependability of LLM applications within this sector.

\begin{figure*}[t]
\centering
\includegraphics[width=1.0\textwidth]
{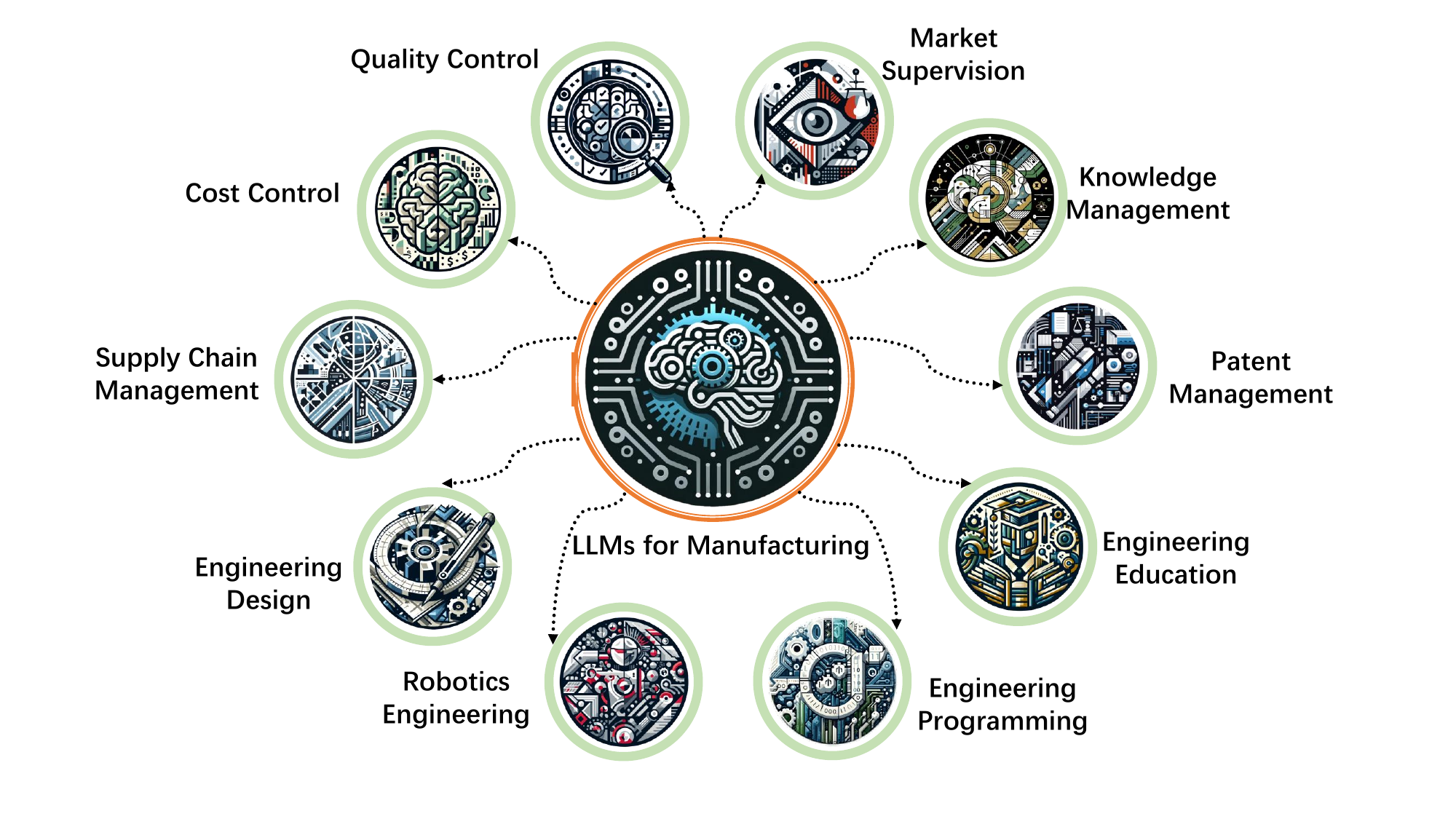}
\caption{LLM applications in manufacturing. Logos in this figure are generated by DALL·E 3.} 
\label{general structure}
\end{figure*}

In summary, while there are inherent challenges in applying LLMs to manufacturing applications (Figure \ref{general structure}), these models are progressively proving to be invaluable tools for data processing, analysis, and generating deep insights and actionable recommendations. Despite the hurdles, continual technological advancements are paving the way for overcoming these difficulties. The future application of LLMs in manufacturing promises to foster significant innovations and opportunities.

\begin{itemize}
\item This paper provides a comprehensive overview of the applications and potential of LLMs in various aspects of manufacturing (Figure \ref{general structure}), including quality control, cost control, supply chain management, engineering design, chip design, robotics engineering, engineering programming, engineering education, patent management, and knowledge management.
\item We evaluate the performance of LLMs (such as GPT-4V) on various manufacturing tasks through examples and case studies, highlighting their strengths in areas like text processing, data analysis, code generation, and zero-shot learning capabilities.
\item This survey identifies the limitations of LLMs in direct computational roles, such as manufacturing design, where they primarily play a supporting role.
\item Our review emphasizes the potential of LLMs to enhance existing manufacturing methodologies and introduce novel strategies in manufacturing applications.
\item We discuss the challenges and future directions for LLM integration in manufacturing, including the need for further refinement, integration with quantitative models, and addressing interpretability and reliability concerns.
\item We underscore the significance of ongoing development and collaboration between academia and industry to fully harness the potential of LLMs in the manufacturing domain.
\end{itemize}

\section{Background}
The manufacturing sector is experiencing a profound shift, fueled by the integration of advanced technologies, evolving market requirements, and an increasing emphasis on efficiency, resilience, and sustainability. This section examines the primary trends and innovations shaping the future of manufacturing, such as the consequences of the COVID-19 pandemic, the emergence of smart factories, the growing focus on workforce development, and the potential of technologies like generative AI.

\subsection{The National Strategy for Advanced Manufacturing}
In October 2022, the Subcommittee on Advanced Manufacturing, part of the National Science and Technology Council's Committee on Technology, released the \textit{National Strategy for Advanced Manufacturing}~\cite{council2022national}. This strategy presents a roadmap for enhancing U.S. leadership in advanced manufacturing to drive economic growth, generate high-quality jobs, promote sustainability, combat climate change, bolster supply chain resilience, ensure national security, and improve healthcare outcomes.
The strategy outlines three key goals to realize this vision:
\begin{enumerate}
\item \textbf{Advance and implement cutting-edge manufacturing technologies}
\item \textbf{Expand and strengthen the advanced manufacturing workforce}
\item \textbf{Enhance resilience across manufacturing supply chains}
\end{enumerate}

For each goal, the strategy outlines several strategic objectives and offers recommendations to be pursued over the next four years. Key areas of focus include enabling clean and sustainable manufacturing, accelerating manufacturing innovation in microelectronics, bio-economy, materials, and smart manufacturing, expanding and diversifying the manufacturing talent pool, developing advanced manufacturing education and training, enhancing supply chain interconnections and reducing vulnerabilities, and strengthening advanced manufacturing ecosystems.

The strategy emphasizes the importance of public-private partnerships, supporting small and medium manufacturers, boosting technology transition, and providing greater access and alignment of federal programs to underserved communities. The strategy aims to enhance coordination across U.S. government agencies and offer long-term direction for federal efforts that promote manufacturing competitiveness.

This national strategy provides a comprehensive framework for advancing U.S. manufacturing capabilities and competitiveness. It recognizes the critical role of technology and innovation, including the potential of emerging technologies such as AI and machine learning, in driving the future of manufacturing. The strategy's emphasis on advancing manufacturing technologies, expanding the workforce, and strengthening resilient supply chains aligns with the critical areas where LLMs and generative AI can be utilized to transform manufacturing processes, as discussed in this paper.

\subsection{The Rise of Smart Factories and Industry 4.0 and 5.0}
Smart factories, powered by Industry 4.0 and 5.0 technologies such as the Internet of Things (IoT), AI, robotics, and cloud computing, are revolutionizing manufacturing processes~\cite{SOORI2023192,rane2023chatgpt}. By integrating these technologies, manufacturers can achieve greater visibility, control, and optimization across their operations, from product design and development to production, quality control, and maintenance. Smart factories enable real-time data collection, analysis, and decision-making, allowing manufacturers to improve efficiency, reduce waste, and respond more quickly to changing market demands. The transition to smart manufacturing is expected to continue, as companies seek to gain a competitive edge through digitalization and automation~\cite{ghobakhloo2018future}.

\subsection{The Growing Importance of Talent Acquisition and Development}
As manufacturing becomes increasingly technology-driven, the need for skilled talent is more critical than ever~\cite{ahmad2024enhancing, betker2023improving}. However, many manufacturers face significant challenges in attracting, developing, and retaining the right talent to support their digital transformation initiatives. To address these challenges, companies are adopting strategies such as partnering with educational institutions to develop industry-relevant curricula, leveraging digital tools for talent acquisition and onboarding, and investing in continuous learning and upskilling programs for their employees. By prioritizing talent development, manufacturers can build a future-ready workforce capable of thriving in the era of smart manufacturing.

\subsection{The Post-Pandemic Manufacturing Landscape}
The COVID-19 pandemic has accelerated the digital transformation of the manufacturing sector, forcing companies to adapt to disrupted supply chains, changing consumer behaviors, and new health and safety protocols~\cite{ambrogio2022workforce}. Despite the challenges posed by the pandemic, many manufacturers have demonstrated remarkable resilience and agility, leveraging digital technologies to maintain operations, support remote work, and optimize production processes. As the industry recovers from the pandemic, manufacturers are increasingly focusing on building resilience, flexibility, and adaptability into their operations to better withstand future disruptions. 

\section{GenAI Application Examples in Manufacturing}

Large language models like GPT-4V can provide NLP capabilities for streamlining communication, Latest LLM like GPT-o1-preview~\cite{zhong2024evaluation} has extraordinary ability in solveing difficult problem. documentation, and knowledge management across distributed teams and supply chains. These models can automate the generation of reports, translate technical documents, and facilitate real-time communication in multiple languages, thereby enhancing collaboration and efficiency. Furthermore, GenAI can analyze vast amounts of data to identify patterns and insights that can inform decision-making, improve forecasting, and optimize production schedules. By integrating these advanced AI capabilities, manufacturers can better manage supply chain complexities, respond swiftly to market changes, and maintain continuity in the face of disruptions.

\begin{itemize}
\item \textbf{Significant application of GenAI for developing and implementing advanced manufacturing technologies:}  Under the first goal, the strategy emphasizes the development and implementation of advanced manufacturing technologies. Generative AI, particularly large language models, can play a significant role in enabling clean and sustainable manufacturing by optimizing processes, reducing waste, and improving energy efficiency. LLMs can analyze vast amounts of data to identify patterns and insights that can inform more sustainable manufacturing practices. Additionally, various AI models can accelerate innovation in areas such as microelectronics~\cite{10440162,du2023generative}, bioeconomy~\cite{al2023inverse,colombeopportunities,pandey2024bioprocessing}, and smart manufacturing by facilitating the design and simulation of new products and processes~\cite{yang2024generative,rane2023chatgpt,rane2024intelligent,doanh2023generative}.

\item  \textbf{Significant application of GenAI for growing the advanced manufacturing workforce:} LLMs can be leveraged to develop personalized learning experiences and training programs that adapt to individual learners' needs and preferences. Transformer-based~\cite{NIPS2017_3f5ee243,liu2023summary} and diffusion-based~\cite{rombach2022high,gozalo2023chatgpt} GenAI can also assist in creating engaging and interactive educational content, such as virtual simulations~\cite{qadir2023engineering,ruiz2023empowering,kaur2023review,li2024aldm,rashid2023geolocation} and intelligent tutoring systems, to attract and retain a diverse talent pool in manufacturing. Furthermore, LLMs can help match job seekers with relevant opportunities and provide career guidance based on their skills and interests.

\item \textbf{Significant applications of GenAI for building resilience into manufacturing supply chains:} Generative AI models can be employed to enhance supply chain visibility and risk management by processing and analyzing large volumes of data from various sources~\cite{EasyChair:12925,stufano2024esplorare,rane2024intelligent,EasyChair:12928}, such as supplier reports, market trends, and geopolitical events. GenAI can help identify potential disruptions and recommend proactive measures to mitigate risks. Moreover, LLMs can facilitate collaboration and information sharing among supply chain stakeholders by enabling natural language interfaces and intelligent decision support systems.
\item  \textbf{Significant applications of GenAI for smart factories and industry 4.0:} In a smart manufacturing environment, language models, multimodal models, and vision models can enhance operational efficiency and product quality. IoT devices gather diverse data types from equipment and production lines, which are processed locally through edge computing. Multimodal models, which combine data from text, audio, and visual inputs, alongside vision models, analyze visual data to identify defects or operational anomalies in real-time. Language models can interpret maintenance logs, operator instructions, and other textual data to provide actionable insights and automated reporting. This integrated approach can provide a seamless flow of information, enabling quicker decision-making and more refined control over manufacturing processes. Large language models can also assist HR departments by automating the screening and initial interviewing processes. 
\end{itemize}

\section{LLMs in Product Development}

\subsection{LLMs for Advanced CAD and CAM}
The integration of LLMs into smart manufacturing design marks a revolutionary advancement, addressing long-standing challenges within the industry. Traditional methods often struggled with scalability, incurred high data acquisition costs, and yielded suboptimal outcomes. In contrast, LLMs stand out by delivering robust performance through their capacity for generalization and their adeptness in handling multimodal, multitask learning processes that draw on expansive datasets~\cite{zhang2023large}. More importantly, the advanced manufacturing revolutionized by LLMs and Artificial General Intelligence (AGI), particularly in chip and Integrated Circuit (IC) design, accelerates manufacturing for microelectronics and semiconductors. This enhances computing hardware, improving both power efficiency and performance. Furthermore, these advancements can be reinvested into the development of LLMs and AGI, creating a positive feedback loop~\cite{zhao2023brain}. It can be foreseen that the integration of LLMs in manufacturing design promotes decarbonization and paves the way for smart manufacturing, aligning technological progress with environmental sustainability.

In the following, we summarize three key innovations brought about by the integration of LLMs into manufacturing design, specifically computer-aided design (CAD) and computer-aided manufacturing (CAM):

\begin{itemize}
    \item \textbf{Data Generation and Generalization Enhancement}: Notable for their exceptional generalization across diverse domains, which stems from the vast and diverse pre-trained datasets, LLMs are able to iteratively generate high-quality training data. This pivotal capability substantially cuts costs and boosts the performance of deep learning applications in CAD and CAM, thereby increasing the adaptability and efficiency of these systems for various manufacturing operations~\cite{Makatura2024Large}.

    \item \textbf{Text-Grounding 3D Content Generation}: LLMs revolutionize CAD by developing generative models that refine parametric designs, a fundamental aspect of mechanical engineering that involves intricate geometric sketches~\cite{poole2022dreamfusion,wucad,lin2023magic3d}. These models, trained on foundational language models, are fine-tuned to generate and interpret these sketches by using structured string formats that define points and connections, facilitating the creation of detailed and accurate CAD designs.

    \item \textbf{3D World Understanding and Reasoning}: Enhancing LLMs with 3D spatial comprehension transforms interactions within both digital and physical manufacturing environments~\cite{hong20233d}. This integration enables sophisticated spatial reasoning, essential for interpreting, analyzing, and refining 3D models in real-world manufacturing applications.
\end{itemize}

These technological advancements not only streamline and enhance the efficiency of design and manufacturing workflows but also ensure precision and innovation in production directives influenced by LLMs.

\subsection{Bridging the Gap Between Idea, Design and Manufacturing}
Despite the initial hesitance due to LLMs' perceived disconnect from on-ground manufacturing activities, their integration into engineering workflows exemplifies their utility. Acting as a crucial link between idea, design, and production phases, LLMs enhance engineering design by improving efficiency, accuracy, and innovative capacity, thus playing a pivotal role in the seamless transition from idea to design to production.

\subsubsection{Text to Design}

\begin{figure*}[t]
\centering
\includegraphics[width=0.8\textwidth]{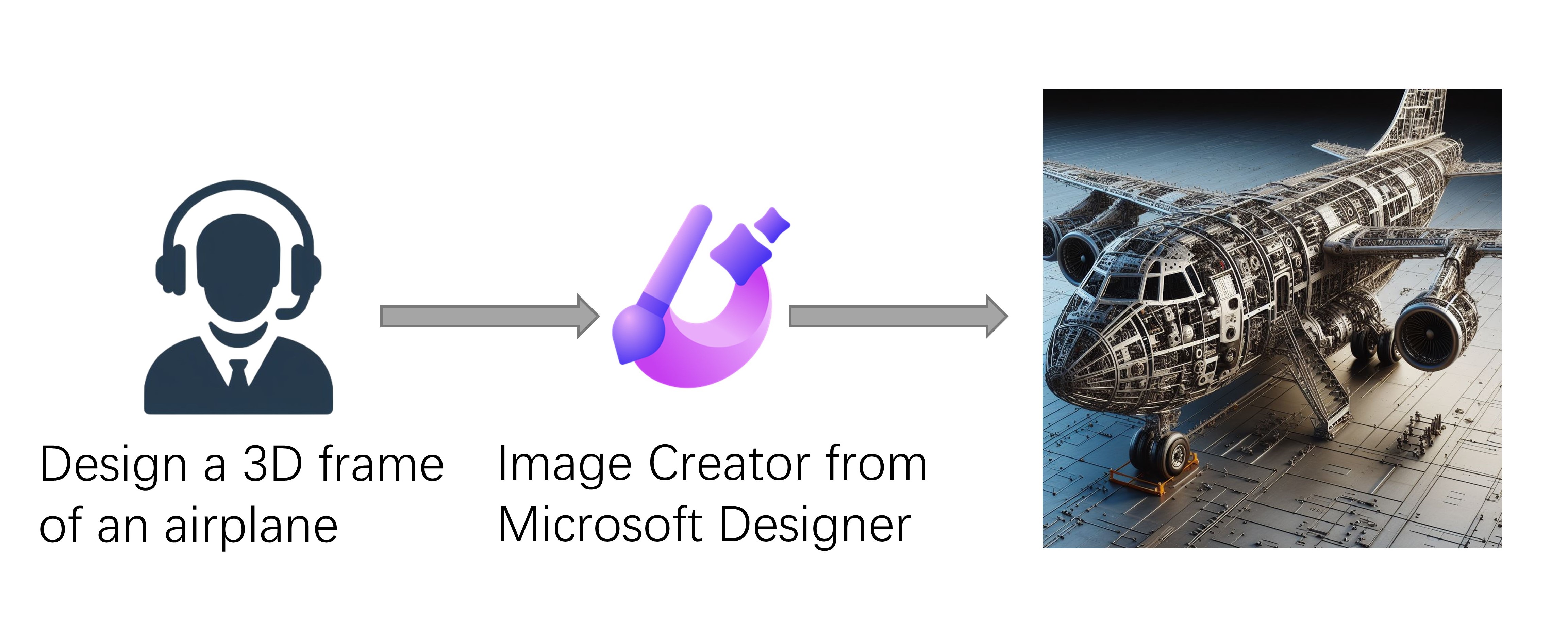}
\caption{Example for Image Creator from Microsoft Designer to generate a 3D frame of an airplane.} 
\label{3d_airplane}
\end{figure*}

\begin{figure*}[t]
\centering
\includegraphics[width=0.8\textwidth]{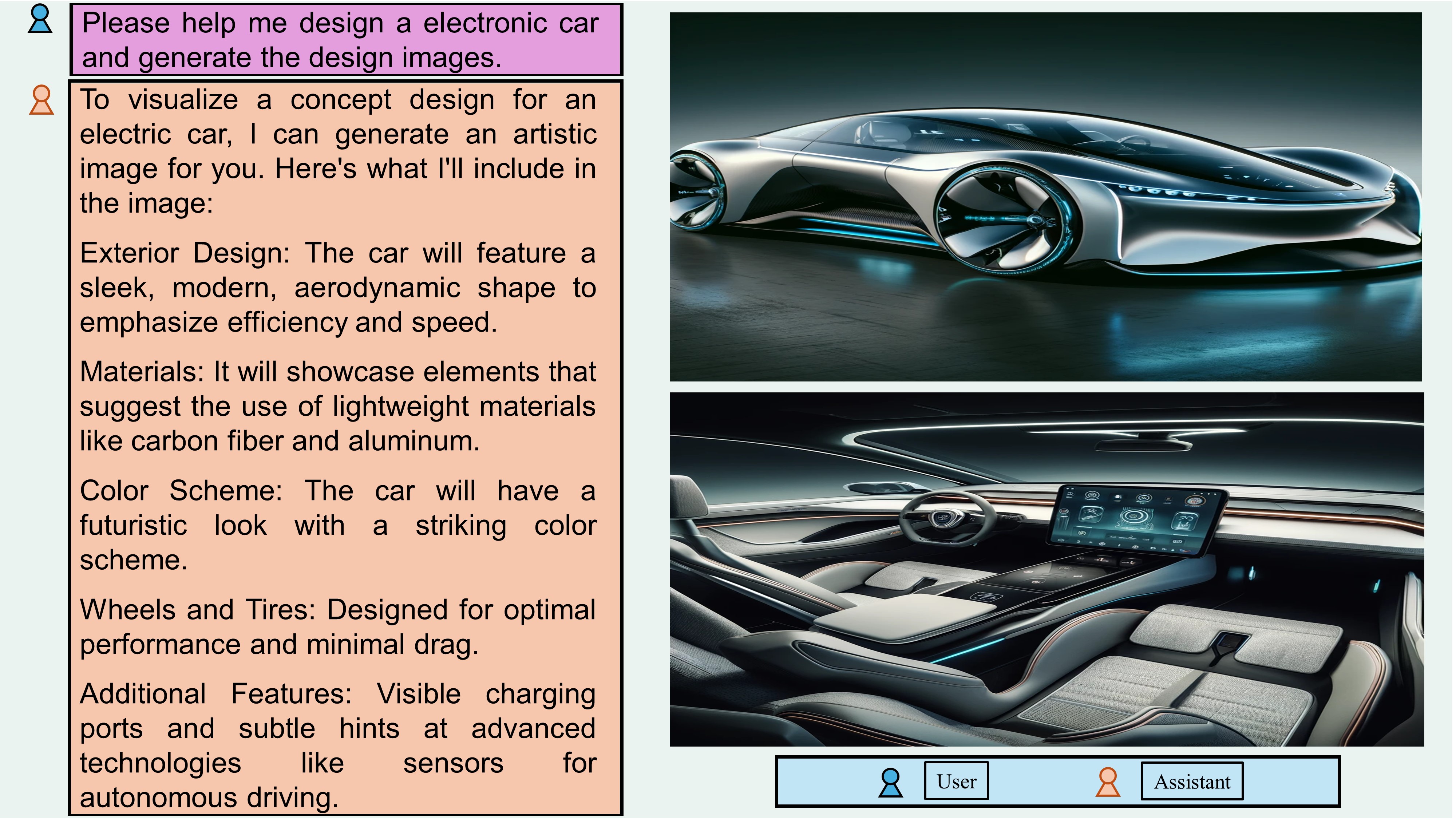}
\caption{Example for LLMs' design of an electronic car. GPT-4V not only gives the key points of advanced car design but also gives possible exterior and interior design drawings of the car.} 
\label{car_example1}
\end{figure*}

The definition of "design" varies widely within the realm of manufacturable~\cite{makatura2023can} outputs and operates across multiple scales. For instance, designing a steel car frame (examples can be seen in Figure \ref{3d_airplane} and Figure \ref{car_example1}) involves a defined design space and specific performance targets, whereas conceptualizing an energy absorption structure, such as a bumper, from an ambiguous description lacks a clear initial mental model. This section explores how LLMs contribute to innovations from both these perspectives.
Traditionally, human-centric design processes have required extensive trial and error, with the visualization of prototypes playing a pivotal role in early assessments. Such visualization is integral to the design process~\cite{preidelKNOWLEDGEENGINEERINGDESIGN2018}, particularly when design parameters are limited. Understanding the design space is crucial for achieving optimal results. The initial step involves visualizing this space; for example, as illustrated below, a steel frame design begins without needing a physical prototype—utilizing open-source formats like SVG to rapidly construct models based on verbal descriptions. Conversely, for more nebulous concepts, LLMs can generate preliminary visuals to spark creativity. Additionally, workflow enhancement tools such as ComfyUI, which integrates stable diffusion, facilitate the exploration of creative concepts, such as adapting anime character elements to automotive design. Advances in technology now allow for the quick creation of 3D designs from 2D inputs~\cite{TripoSR2024}, signaling a promising future where current 2D images, like those generated by  DALL·E3, still fall short of providing adequate spatial data for 3D modeling.

\subsubsection{Design to Manufacturing}

\textbf{Computer-Aided Design (CAD)}, exemplified here~\cite{fuGPT4AIGChipNextGenerationAI2023,naveed2023comprehensive}, showcases how Generative Pre-trained Transformers (GPTs)~\cite{he2024chateda} significantly enhance coding within this realm. Present-day CAD software typically features APIs that facilitate automated model construction. Language models interpret textual inputs to generate initial design drafts~\cite{picard2023concept}, accelerating the preliminary phases of the CAD process. This function promotes rapid design iterations and conceptual refinements prior to detailed CAD development. Moreover, these models streamline the CAD workflow by automating mundane tasks, recommending design adjustments based on established best practices, and flawlessly integrating intricate design specifications. They also optimize the design-to-manufacturing sequence, ensuring that outputs are both functional and manufacturable, thereby minimizing extensive redesigns and prototype development~\cite{li2019artificial,ahmad2024enhancing}. This application can be succinctly termed GPT-CAD.

\begin{figure*}[t]
\centering
\includegraphics[width=1.0\textwidth]{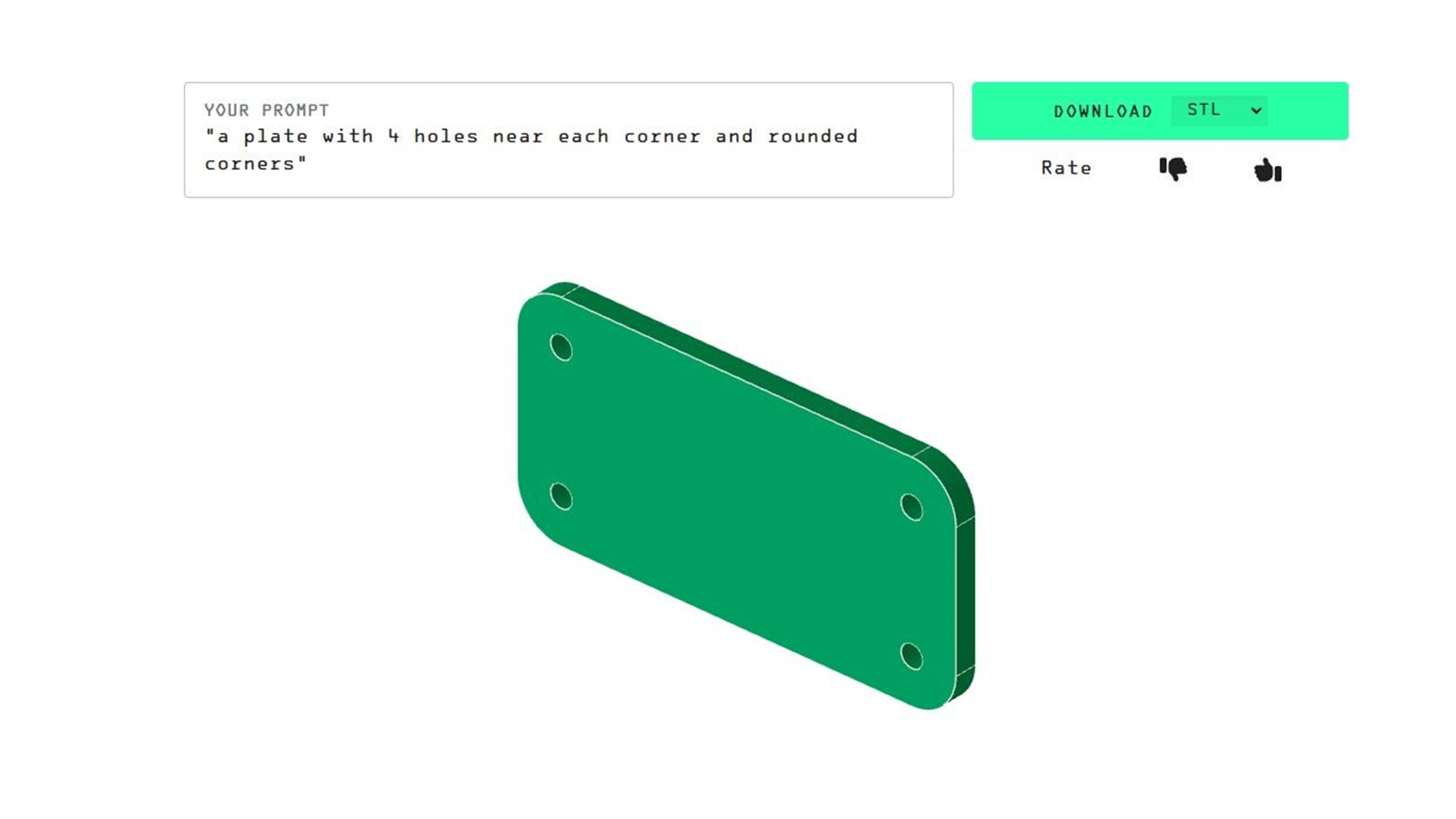}
\caption{Example for Text-to-CAD to generate a plate with 4 holes near each corner and rounded corners using ASCII STL format. Text-to-CAD is an open-source prompt interface for generating CAD files through text prompts. The infrastructure behind Text-to-CAD utilizes ZOO Design API and Machine Learning API to programmatically analyze training data and generate CAD files~\cite{zoo_dev_text_to_cad}.} 
\label{CAD_example}
\end{figure*}

For instance, upon receiving a command such as "create a 3D model of a steel frame" GPT-CAD interprets this and translates it into a sequence of FreeCAD operations to construct the specified geometric model (example can be seen in Figure \ref{CAD_example}). GPT-CAD shows promising educational applications; it significantly lowers language and technical hurdles, contributing to more inclusive and adaptive learning environments. By responding to natural language inputs, GPT-CAD adapts to various cognitive and linguistic preferences, fostering a more personalized educational experience. This adaptability is particularly beneficial for students with disabilities or those who might find the technical complexity of conventional CAD interfaces daunting. Beyond educational applications, GPT-CAD proves invaluable in creative design, facilitating swift parametric design alterations. This advancement may redefine the role of designers, transitioning them from sole creators to orchestrators of AI-assisted design processes~\cite{dhar2024can,machadoParametricCADModeling2019,kapsalis2024cadgpt}.

\begin{figure}[t]
    \centering
    \includegraphics[width=0.75\linewidth]{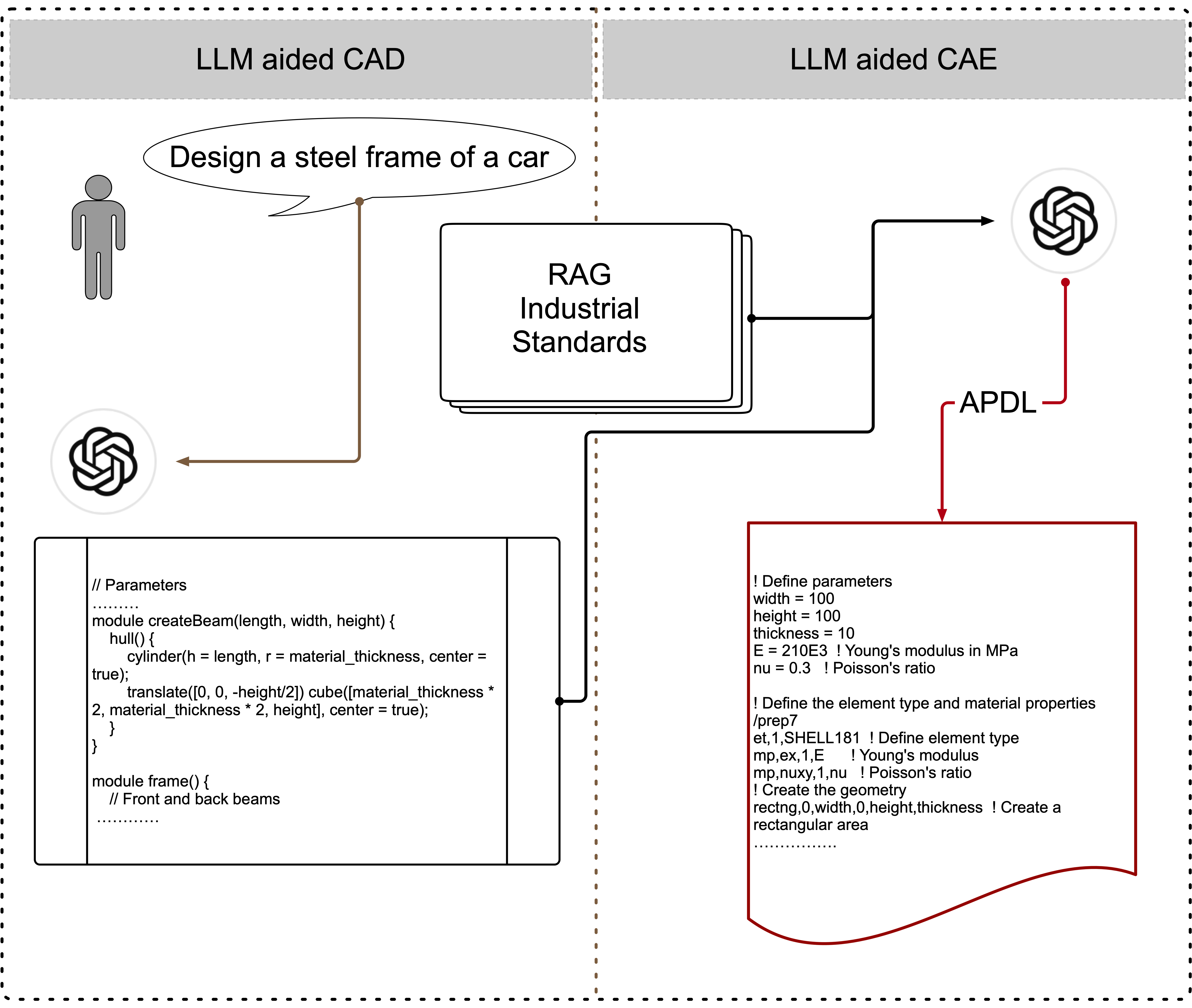}
    \caption{With the knowledge base built with RAG, LLMs can help researchers build models that can do designs and experiments efficiently.}
    \label{CAD2}
\end{figure}

\textbf{Computer-Aided Equipments (CAE)}, as referenced in recent studies~\cite{nelsonUtilizingChatGPTAssist2023,yangFLUIDGPTFastLearning2023}, are integral to manufacturing processes, wherein GPTs significantly contribute by facilitating and conducting simulations due to their superior coding capabilities (example can be seen in Figure \ref{CAD2}). Additionally, the implementation of Retrieval-Augmented Generation (RAG)~\cite{jiang2023active,mao2020generation,gao2023retrieval} enables GPT-CAE to serve educational purposes and assist researchers in comprehending industrial standards. In industries such as automotive or construction, the ability to easily access data on intersection shapes or the mechanical properties of materials proves beneficial; thus, GPT-CAE acts as both an assistant and a resource repository. Multimodal LLMs, like the state-of-the-art GPT-4V, facilitate the analysis of simulation results. For instance, in mechanical engineering, identifying maximum stress points and analyzing stress contours are crucial for understanding potential failures and optimizing designs. GPT-4V can pinpoint potential failure locations and conduct analyses. Due to the high cost and current unavailability of fine-tuning for the latest models from OpenAI, combined approaches using computer vision and LLMs, such as employing YOLO with GPT-3.5 for detection and pattern analysis, are also utilized. Discussing CAE leads to the consideration of real experiments, which similarly benefit from standardized processes and data collection~\cite{liu2023llm}.

One of the primary objectives of CAE is optimization, which relies on simulation results and defined optimization goals. GPTs can play a pivotal role in this context by providing inspiration and constructing optimization target index functions. With GPT assistance, a variety of strategies can be presented as a toolbox, allowing the GPT to conduct simulations. Based on the outcomes, LLMs can act as agents, strategizing subsequent steps and thereby relieving human experts from the burden of simulation management, allowing them to focus on theoretical advancements. Due to the advanced capabilities of GPT-4V, it can serve as a supervisor in reinforcement learning, a popular method in serial optimization processes.

\subsection{Application: Multi-agent Frameworks}
Current pre-trained models, though advanced, are insufficiently autonomous to complete complex tasks solely through self-reflection. Multi-agent frameworks represent a strategy for enhancing the performance of next-generation AI. The core concept involves treating multiple machine learning models as team members, which necessitates a well-established information-sharing protocol. While LLMs are not essential in this setup, we still utilize GPT-4V as agents within the team.

\begin{figure*}[t]
\centering
\includegraphics[width=0.8\textwidth]{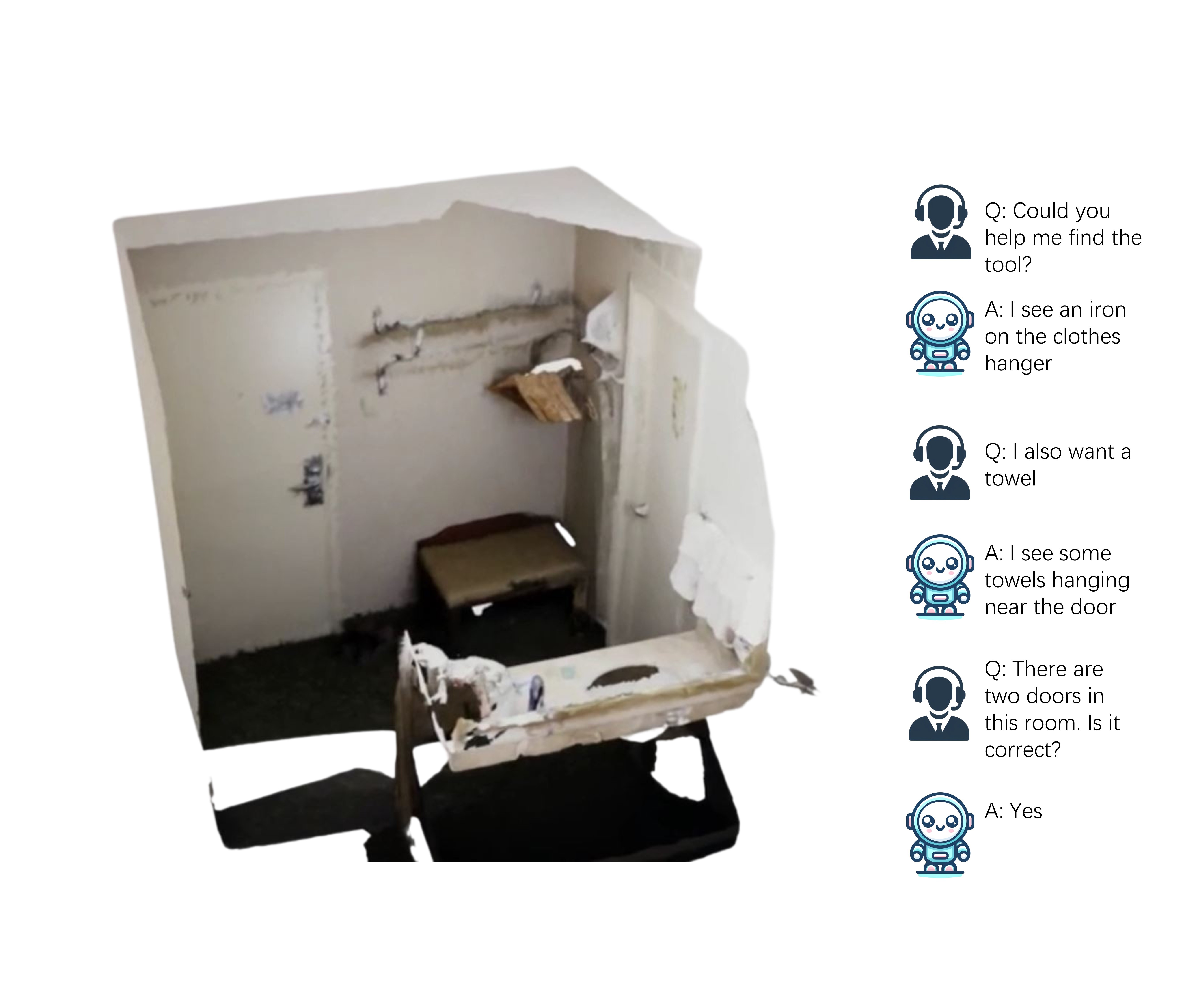}
\caption{Example for 3D-LLM in 3D world understanding and question answering. Logos in this figure are generated by DALL·E 3.} 
\label{3d_world_example}
\end{figure*}

\textbf{Human assistant} strategies rooted in the agent concept are gaining traction, especially in the manufacturing and engineering sectors, where specific tasks are delegated. For example, AI-driven sensors~\cite{rayAISensorResearch2023} enhance the intelligence of sensing systems, which play a vital role in collecting information during manufacturing processes (as illustrated in Figure \ref{3d_world_example}). By combining AI's computational power with human intuition and expertise, researchers can achieve greater precision, efficiency, and creativity in their sensor-based studies. This collaboration between AI and human insights creates a dynamic research environment, driving sensor research forward and opening new pathways for exploration. In this setting, GPT-4V serves as a crucial virtual assistant, aiding in information retrieval, proposing methodologies, and identifying potential challenges. Its predictive capabilities enable researchers to foresee and address obstacles effectively, saving both time and resources while ensuring strong research outcomes.

\begin{figure}[t]
    \centering
    \includegraphics[width=0.75\linewidth]{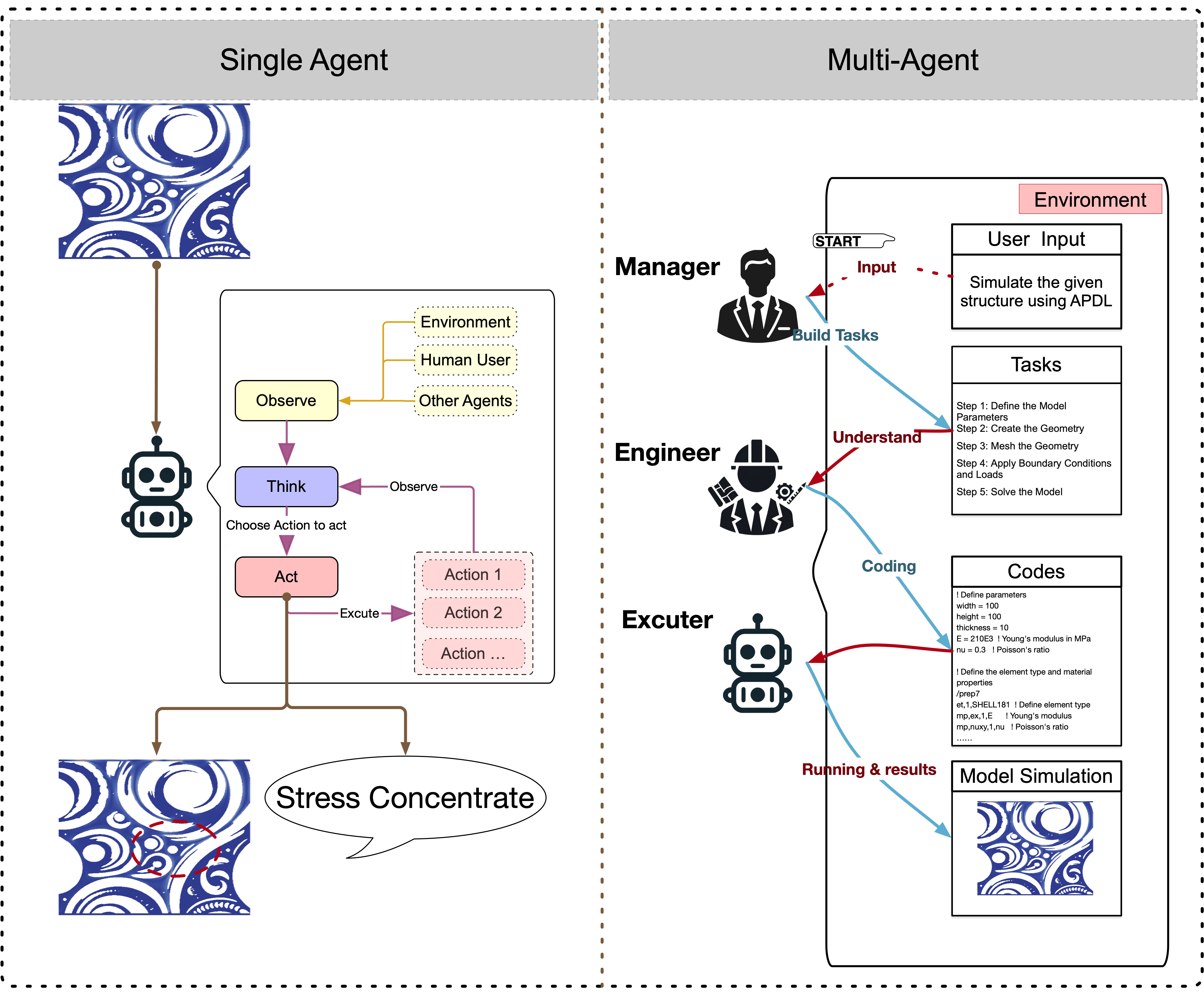}
    \caption{Example of agent and multi-agent framework.}
    \label{Multi-agent2}
\end{figure}

\textbf{Optimization} and similar tasks are often highlighted as beneficial applications of the multi-agent framework~\cite{wang2024visiongpt}. Typically, a multi-agent team comprises a manager, an engineer, and an executor (example can be seen in Figure \ref{Multi-agent2}). The manager divides the task into manageable segments, the engineer then devises solutions for each segment—such as writing code—and the executor implements these solutions, for instance by executing the code to construct a model in FreeCAD. Consider the design of a desk: the manager specifies components like the tabletop and legs, including their dimensions and positions, while the engineer provides the necessary coding for the model, and the executor subsequently uses FreeCAD to construct the model.

\subsection{Examples of LLMs' Application in Production Development}
\subsubsection{Automobile Design}

\begin{figure*}[t]
\centering
\includegraphics[width=0.75\textwidth]{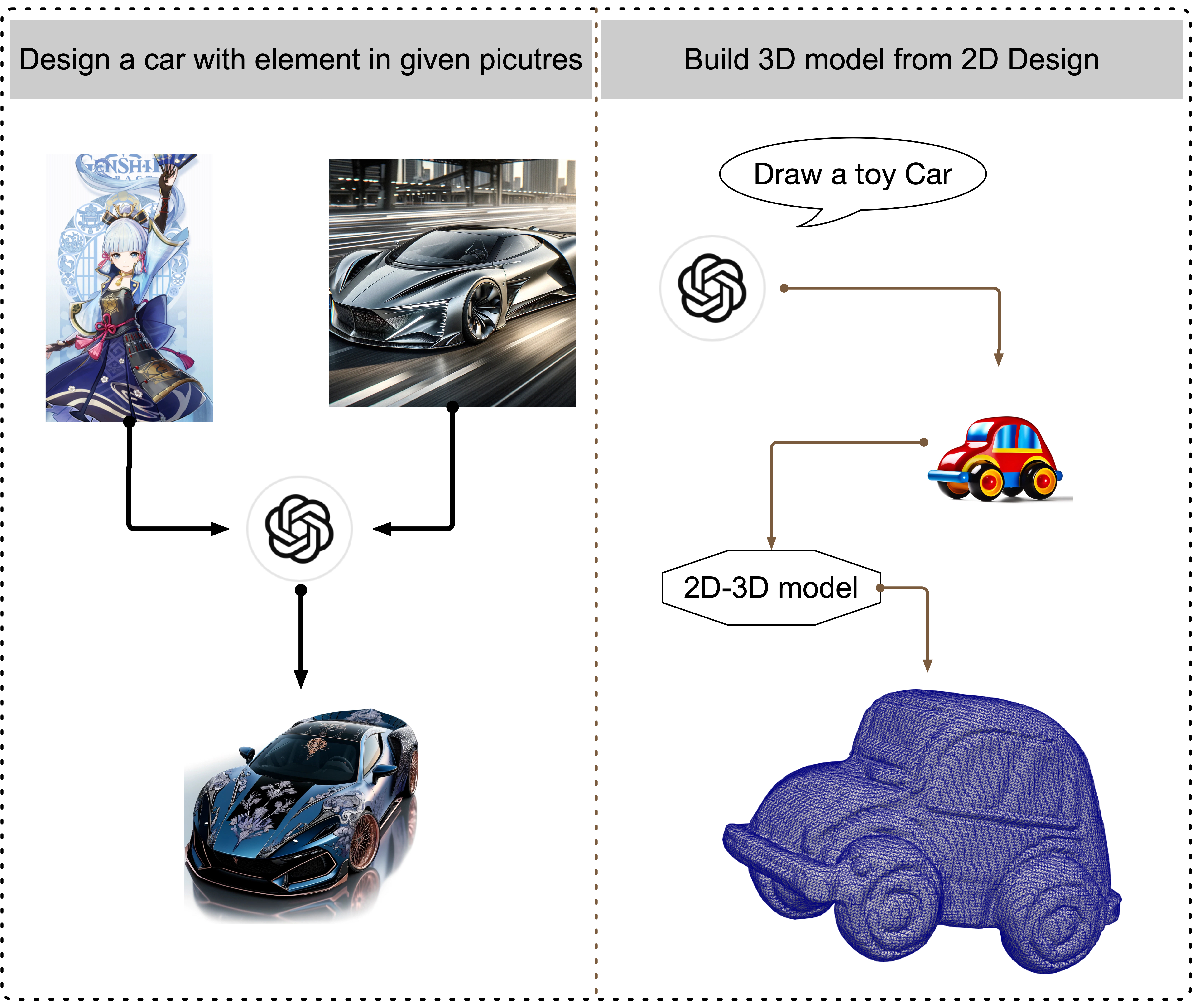}
\caption{Left: An example of text design based on GPT-4V, extracting elements from the input to aid in the design of a sports car's appearance. Right: A straightforward generation from GPT-4V possesses accurate spatial information suitable for creating a 3D model.} 
\label{car_example2}
\end{figure*}

\begin{figure*}[t]
\centering
\includegraphics[width=0.8\textwidth]{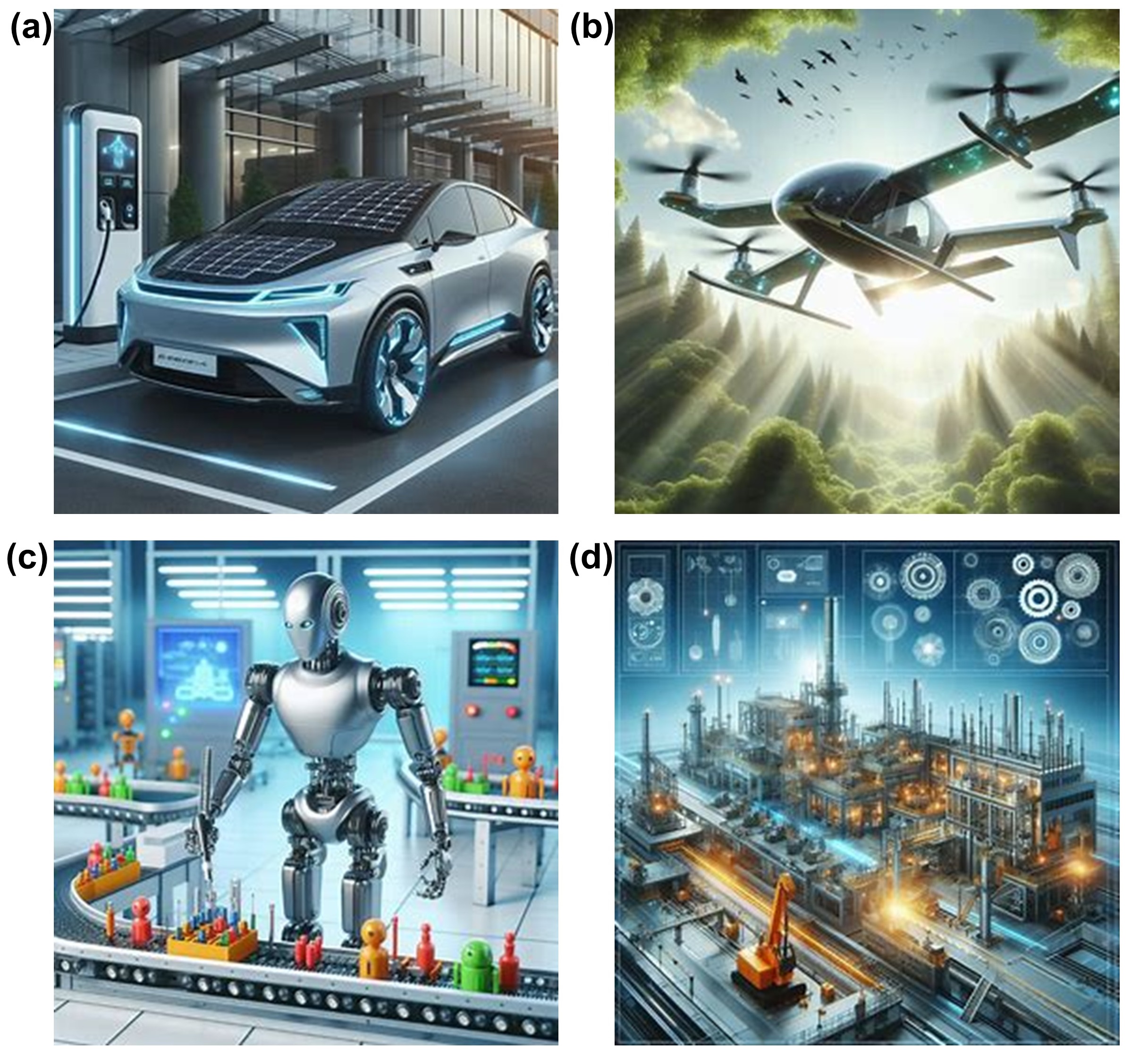}
\caption{Example for  DALL·E3' s ability to design engineering products. a) Design of an electrical car. b) Design of an electrical plane. c) Design of a manufacturing robot. d) Design of a manufacturing pipeline.} 
\label{3d_world_example1}
\end{figure*}

LLMs offer substantial support in the domain of automobile design, as elucidated through a multi-faceted approach (example can be seen in Figure \ref{car_example2} and Figure \ref{3d_world_example1}):
\begin{itemize}
    \item Innovative Design Conceptualization: LLMs possess the capability to dissect and analyze historical data, prevailing trends, and consumer preferences. This analytical prowess aids designers in the generation of innovative automobile design concepts. By engaging with LLMs, designers can acquire insights into future car design trends, explore the potential for utilizing novel materials, and consider the incorporation of eco-friendly technologies.
    \item Enhanced Decision-Making Through NLP: LLMs excel in understanding and processing a myriad of technical documents and communications emanating from diverse design and engineering teams. This facilitates informed decision-making among team members, enabling them to judiciously balance considerations such as vehicle safety, energy efficiency, and cost-effectiveness.
    \item Document Workflow Automation: The automobile design process is characterized by the generation of extensive documentation and specifications. LLMs can automate the processes of generation, review, and updating of these documents, thereby reducing manual errors and enhancing operational efficiency.
    \item Simulation of Customer Feedback and Market Analysis: By analyzing copious amounts of user reviews and market research data, LLMs can simulate potential customer reactions to specific designs. This simulation assists designers in refining their products to align with consumer expectations prior to market launch.
    \item Educational and Training Utility: LLMs serve as an invaluable educational resource, facilitating the dissemination of cutting-edge automotive design concepts, tools, and techniques. Through interactive learning modalities, practitioners are afforded opportunities for continual professional development.
    \item Technical Support and Clarification: In instances where designers and engineers confront specific technical challenges, LLMs provide immediate support and clarification, aiding in the comprehension of complex engineering principles and the resolution of technical issues.
\end{itemize}

These capabilities render LLMs a formidable tool within the automotive design sphere, enhancing both the innovation and efficiency of design processes while enabling companies to more effectively meet market demands and consumer expectations.
\subsubsection{Aerospace Design}
The introduction of LLMs in aerospace can help reduce the time needed to produce models used in manufacturing and design. Traditional optimization processes involve creating a Computer-Aided Design (CAD) model, running simulations of the design, analyzing the results to determine changes that need to be made, and updating the model to reflect the changes. The process is repeated until the results produce no noticeable improvements. In the traditional process, one of the largest time sinks is the fixing of the CAD model to reflect the changes made in each step. The ability to quickly generate designs has the potential to change how the optimization process of an aircraft is performed. Through the use of LLMs, the turnover time from iteration to iteration can potentially be drastically reduced. In addition to potential time savings in updating the CAD model throughout the process, the ability of LLMs to quickly analyze data can help improve the manufacturing/optimization process. Computational Fluid Dynamic (CFD) simulations output a lot of data that is used to judge the performance of the current design and inform changes that need to be made for future iterations. The proven ability of ChatGPT-4o to analyze data could potentially speed up the data analysis aspect of the manufacturing and design process.

\begin{figure*}[t]
\centering
\includegraphics[width=0.8\textwidth]{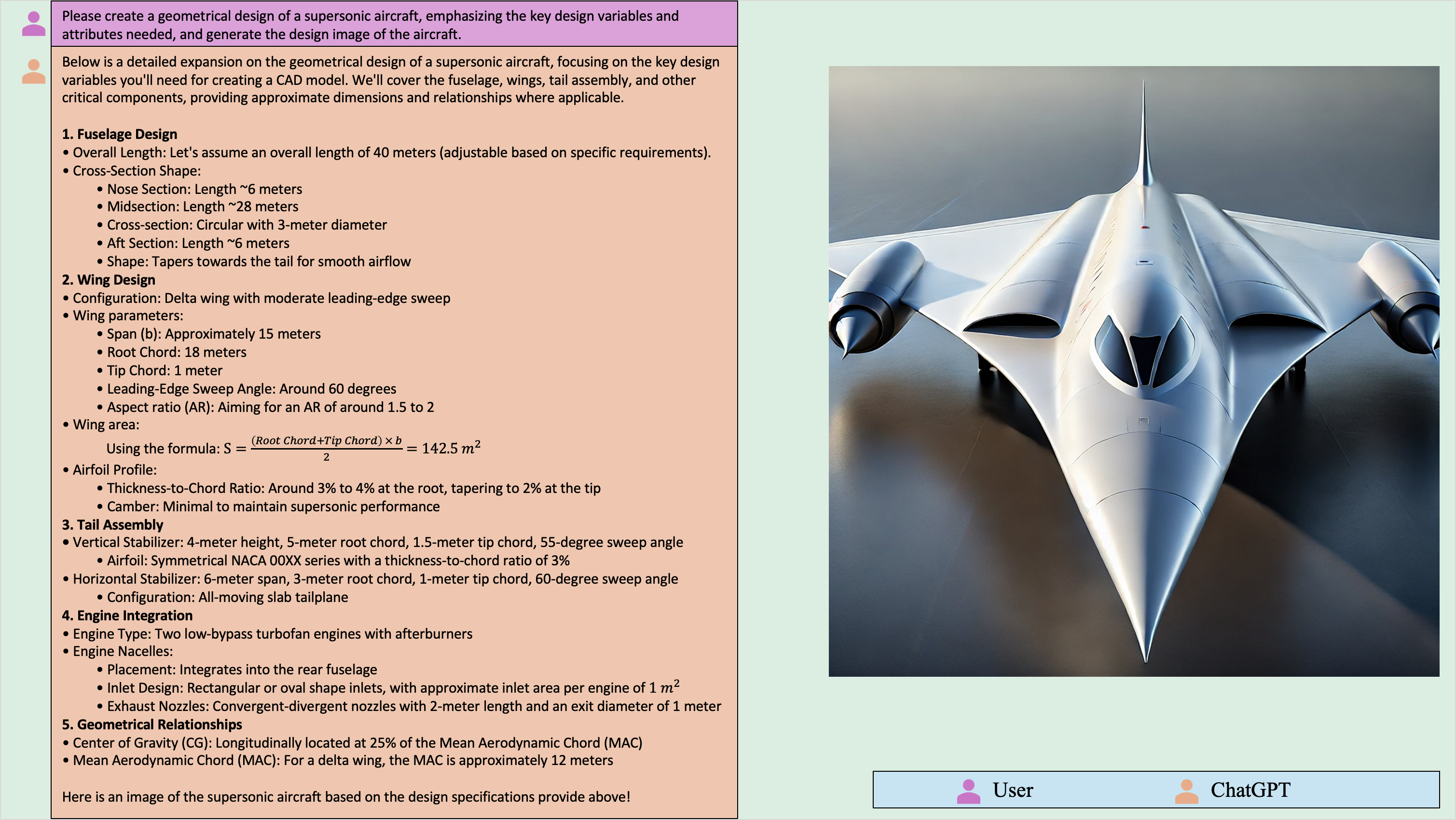}
\caption{Example of ChatGPT-4o's ability to create an image of a supersonic aircraft using specific design variables from a user prompt.} 
\label{Aerospace}
\end{figure*}

Using LLMs, the overall design process for any aerospace vehicle can be made through a few simple prompts; this ranges from subsonic aircraft to supersonic to rockets and much more. By asking for a general design and the key variables for a supersonic aircraft, ChatGPT-4o was able to provide a step-by-step list of what should be included in the design and why it was making those specific design choices. The LLM was able to utilize key aerospace engineering equations with already provided design variables to create a functioning design at supersonic speeds, such as blended-body and integrated engines into the aircraft. It was then able to generate what this aircraft model should look like based on (example can be seen in Figure \ref{Aerospace}). The ability of the LLM to recognize the importance of these features indicates an ability to give design ideas for complex projects in the aerospace industry, ultimately streamlining and improving the aerospace design process by rapidly generating new and innovative ideas.

\subsubsection{Chip Design and Manufacturing}
Over the last few decades, the evolution of chip design and manufacturing has become a cornerstone of modern electronic technology, involving the integration of electronic components into chips and the translation of these complex designs into practical manufacturing processes using silicon and other materials. As electronic technologies rapidly advance, the complexity and demands of chip design and manufacturing have dramatically increased. According to the address in the research report from President’s Council of Advisors on Science and Technology (PCAST)~\cite{President’sCouncilofAdvisorsonScienceandTechnology_2024}, in response to escalating demands in semiconductor design, the application of AI has become instrumental in enhancing the efficiency and quality of advanced chip production. Broad utilization of AI techniques has facilitated significant reductions in the design time and engineering resources necessary for developing state-of-the-art semiconductors, which are pivotal both economically and for national security~\cite{RevitalizingtheU.S.semiconductorecosystem_2022}. This progress in AI applications not only supports the rapid evolution of electronic technologies but also fosters innovation through a more inclusive range of participants in the semiconductor industry, enhancing global competitiveness in design capabilities. This is also particularly evident in the current landscape where the U.S. government's strategic initiatives under the CHIPS Act have catalyzed significant advancements. 

The CHIPS Act, aimed at revitalizing the U.S. economy and enhancing its competitiveness on the global stage post-pandemic, has initiated a remarkable transformation in the semiconductor industry. A substantial investment, highlighted by President Biden's recent announcement of a \$6.1 billion allocation to Micron Technology, underscores this commitment. This funding is part of an expansive effort under the CHIPS Act that has already stimulated \$327 billion in projected investments and a significant increase in domestic semiconductor manufacturing facilities.

By fostering an environment conducive to growth, the Act is set to increase the U.S.'s share of the world's advanced semiconductor production from virtually zero to about 20\% by 2030. This strategic expansion is not only crucial for reducing dependency on global supply chains—a vulnerability starkly highlighted during the pandemic—but also for reinforcing the U.S. position in high-tech industries. 

In this context, the application of LLMs in semiconductor chip design is emerging as a transformative force in accuracy and productivity (example can be seen in Figure \ref{chip_design_example1}). Leveraging advanced deep learning and NLP capabilities, LLMs address the complex tasks and requirements of chip design, offering engineers smarter, more efficient design solutions. The synergy between federal initiatives and cutting-edge technologies like LLMs illustrates a robust pathway forward, ensuring that chip design continues to play a pivotal role in the technological landscape of the future. This alignment highlights not only the increased capacity for domestic manufacturing but also the potential for significant advancements in chip technology and design methodologies, which are crucial for maintaining technological leadership and economic strength.

\subsubsection{LLMs for Biomanufacturing}
\begin{figure*}[t]
\centering
\includegraphics[width=0.8\textwidth]{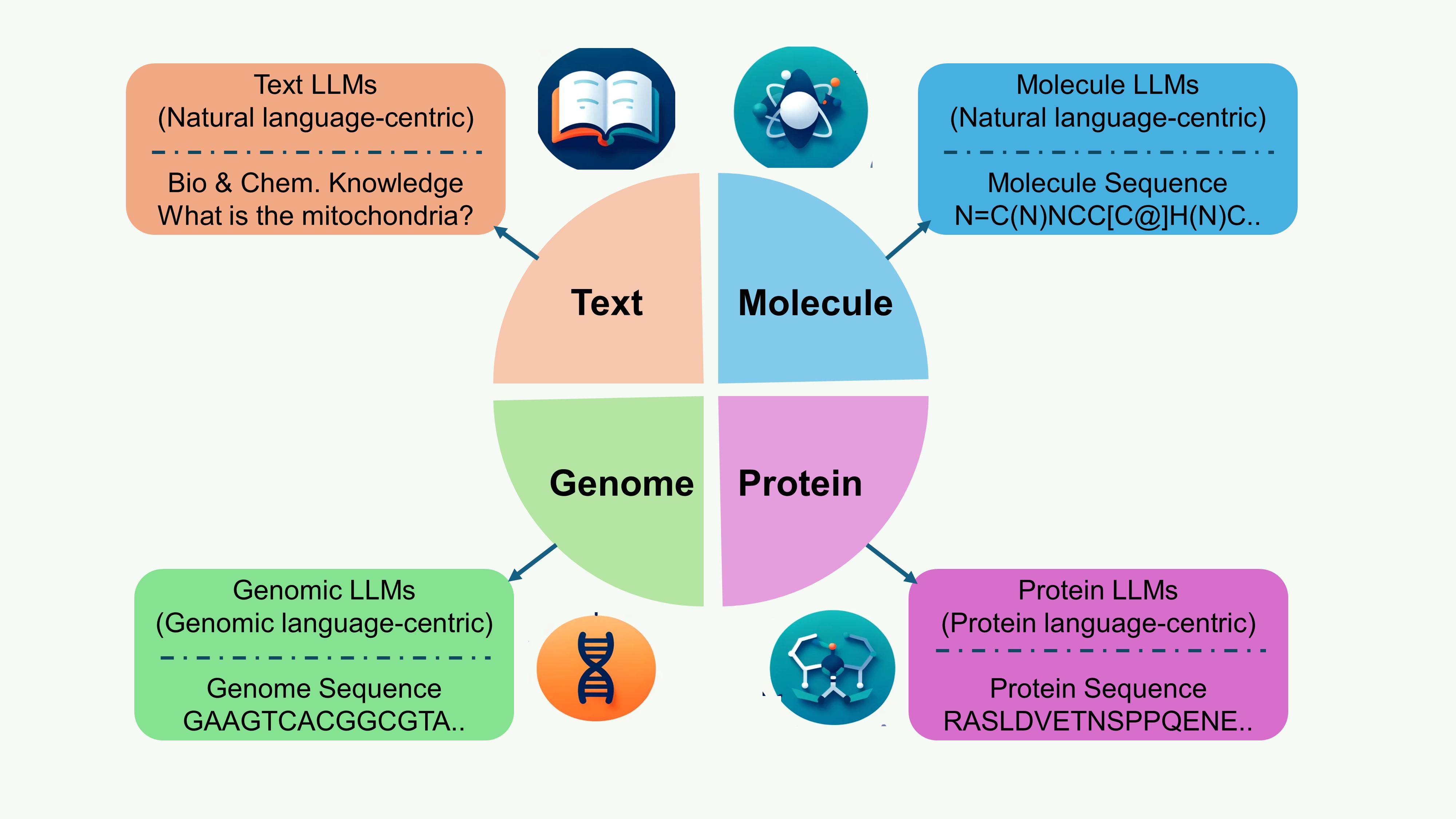}
\caption{LLMs for bio-engineering. LLMs can be utilized in genomic, protein, biological general text, and molecule areas. } 
\label{bio-engineering_example1}
\end{figure*}

\paragraph{LLMs for Bio-Economy}
LLMs play a significant role in advancing the bioeconomy by facilitating research, innovation, and the effective translation of scientific discoveries into practical applications\cite{council2022national}. 
The applications of advanced biotechnologies and biomanufacturing are leading the trends of fulfilling the executive orders for accessibility, safety, and security, in which the LLMs have played the role as the provider for innovative solutions.
This is because LLMs can process vast amounts of scientific literature and data, providing researchers, developers, and manufacturers with insights that accelerate the design and optimization of bio-engineering processes. These models aid in the development of genetic and proteomic engineering tools, support the engineering of multicellular systems, and enhance the predictive modeling capabilities essential for biological processing (Figure \ref{bio-engineering_example1}). Additionally, LLMs can contribute to workforce education and training, making complex biotechnological concepts more accessible and fostering a skilled workforce capable of driving the bioeconomy forward. Through these capabilities, LLMs strengthen the bioeconomy's foundation in sustainability, health, and security

\paragraph{LLMs for Molecules} 
A variety of molecular language models have been developed to predict molecular properties and facilitate the generation of new molecules. These models utilize the chemical language system, such as SMILES annotations, which encode molecular structures as textual strings~\cite{weininger1988smiles}. By representing molecules through SMILES strings, these models can leverage unlabeled data for self-supervised learning, enabling them to comprehend and represent molecular structures without needing explicit task supervision.

Similar to applications in natural language processing, the architecture and pre-training strategies of large language models (LLMs) have been widely adopted in molecular modeling. For example, GPT has significantly impacted molecular generation in cheminformatics. Mol-LLMs, including MolGPT~\cite{bagal2021molgpt}, SMILESGPT~\cite{adilov2021generative}, and Taiga~\cite{wang2022pre}, primarily rely on SMILES strings as input, providing an effective solution for navigating the extensive chemical space. These pre-trained chemical models not only enhance traditional molecular property prediction but also accelerate molecular design and drug discovery efforts.

\subparagraph{Fundamental Chemical Molecular Tasks}
One of the core tasks in computational chemistry is predicting molecular properties. This includes forecasting key characteristics such as solubility, lipophilicity, affinity, absorption, distribution, metabolism, excretion, toxicity, and biological activity, all based on the chemical structure of a molecule. Typically, models in this domain are pre-trained on large, unlabeled datasets and then fine-tuned with more specific data to enhance prediction accuracy. Notable BERT-based models used in this field include ChemBERTa~\cite{chithrananda2020chemberta}, MAT~\cite{maziarka2020molecule}, Mol-BERT~\cite{li2021mol}, and MG-BERT~\cite{zhang2021mg}.

In the molecular biomanufacturing space, predicting drug-drug interactions (DDIs) is another critical task within drug discovery. DDI prediction involves assessing potential interactions that may occur when multiple drugs are administered together. This is vital for understanding how various pharmaceutical compounds could interact in the body, guiding the development of new drugs. X-Mol~\cite{xue2020x} stands out as a prominent model in this area, significantly improving the accuracy and reliability of DDI predictions.

\subparagraph{Molecular Design}
Molecular generation has emerged as a key focus in computational chemistry~\cite{adilov2021generative} and drug design, leveraging advanced computational techniques to create novel molecular structures. This is particularly valuable in the development of new drugs and materials, as the ability to generate molecules with desired properties can accelerate the discovery process. Molecular generation generally follows two approaches: template-based design and de novo design. The template-based method involves starting with an existing molecular framework and making modifications to enhance desirable properties or minimize unwanted effects. In contrast, de novo design creates entirely new molecular structures from scratch, without relying on existing templates. This approach is especially useful for discovering unique compounds with potential therapeutic or material applications. It requires computational models capable of exploring vast chemical spaces and sophisticated algorithms to predict viable molecular structures.

To support these goals, various large molecular language models (Mol-LLMs) have been developed. MolGPT and SMILESGPT, for example, have proven the utility of LLMs in molecular generation. MolGPT, one of the pioneers in using GPT for molecular creation, applies conditional training to optimize molecular properties and is well-regarded for its efficiency in molecular modeling and drug discovery. Its ability to control multiple properties allows for precise molecule generation. SMILESGPT, built on the GPT-2 architecture, leverages a causative Transformer tailored for drug discovery, trained using SMILES notation. Another example is BARTSmiles~\cite{chilingaryan2022bartsmiles}, a powerful generative masked language model designed for molecular tasks. It utilizes extensive self-supervised pre-training on over 1.7 billion molecules, which significantly enhances its performance across various molecular tasks.

There are also research works\cite{liu2024large, kuenneth2023polybert, xu2023transpolymer} aiming for predicting polymer properties by integrating large language models (LLMs), physics-based modeling, and experimental measurements. Demonstrated through polymer flammability metrics, this approach significantly enhances prediction accuracy and enables efficient exploration of the vast chemical space of polymers with minimal experimental data.

These models collectively represent the forefront of computational approaches in molecular generation, highlighting the importance of integrating advanced machine learning techniques with chemical informatics. Their ability to generate novel molecules with tailored properties accelerates the discovery process in drug development and material science, paving the way for innovative solutions in these fields.

\subparagraph{LLMs for Protein Study}
The use of LLMs in protein research primarily centers on three core areas: protein function prediction~\cite{madani2023large}, protein sequence generation, and protein structure prediction. These models play a crucial role in advancing our understanding of protein function, contributing to drug design and various biomedical research efforts~\cite{anishchenko2021novo}.

\begin{figure*}[t]
\centering
\includegraphics[width=0.8\textwidth]{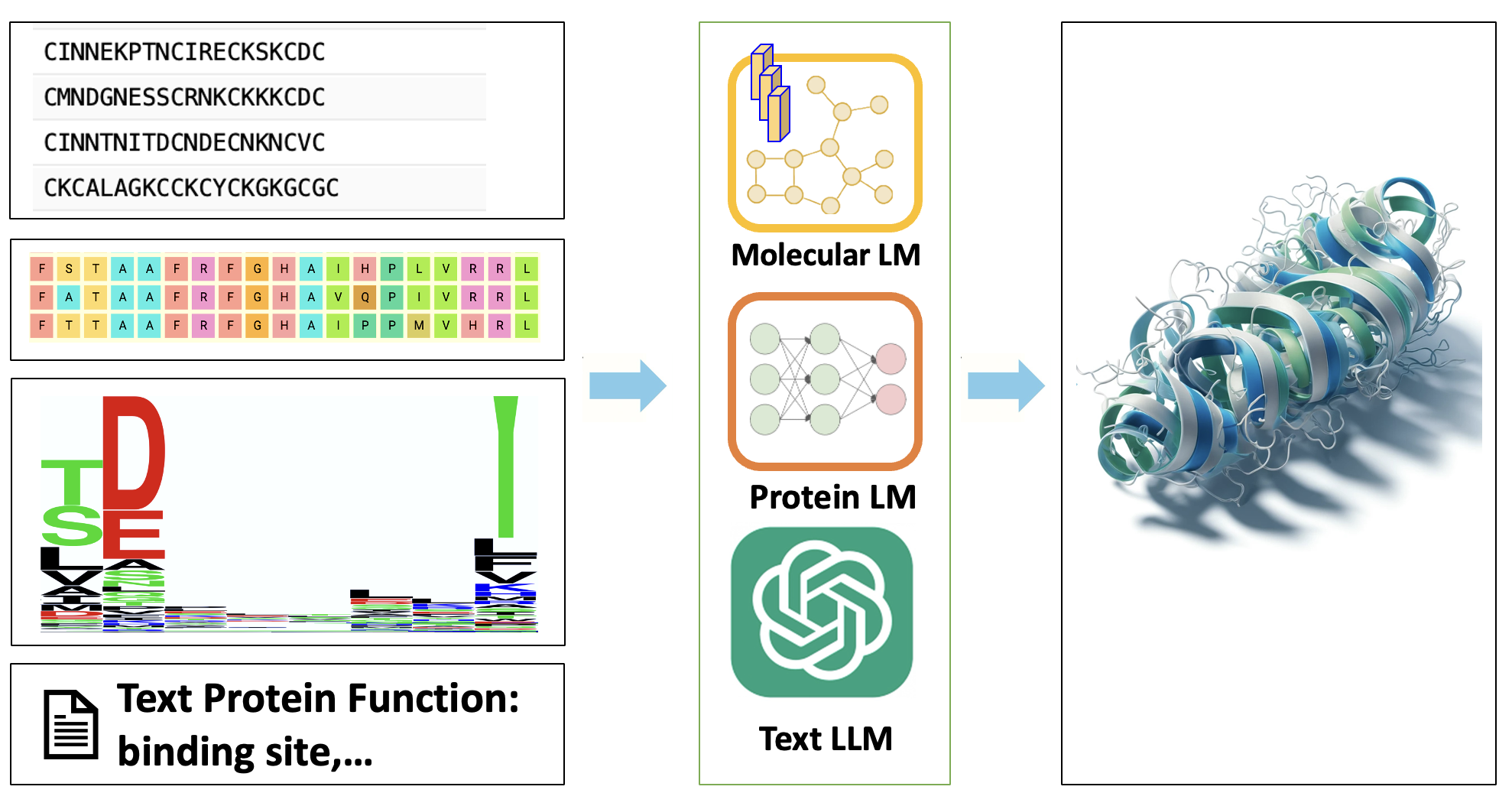}
\caption{Example of protein structure predictions. LLMs hold significant potential for protein design, including generating protein sequences, predicting new structures, and creating protein schematics based on sequence encodings.}
\label{protein_example1}
\end{figure*}

\textbf{Protein structure prediction:} This task involves determining the three-dimensional structure of a protein from its input sequence~\cite{das2021accelerated}, including atomic coordinates and the topological relationships between atoms. Predicting protein structures is essential for understanding how proteins function~\cite{shin2021protein} (tertiary structure prediction) and the operation of complex multi-subunit proteins, such as hemoglobin (quaternary structure prediction). Many of these models leverage encoder-based Prot-LLMs (like the ESM family) to extract information from sequences and perform structure prediction~\cite{huang2022backbone}. Advances in technology have also led to solutions that incorporate diffusion models~\cite{abramson2024accurate}.

\textbf{Protein function prediction:} This area focuses on identifying the biological roles of proteins, including their functions within organisms and interactions with other biomolecules. Functional predictions encompass various tasks, such as protein classification, prediction of protein-protein and protein-ligand interactions, subcellular localization prediction, distant homology detection, fluorescence property prediction, stability prediction, lactamase activity prediction, and solubility prediction, among others.

\textbf{Protein sequence generation:} This process involves generating new amino acid sequences that can potentially form functional proteins for applications ranging from drug design and enzyme engineering to fundamental biological research. Protein sequence generation typically falls into two categories: designing entirely new sequences and optimizing existing ones. Autoregressive models, such as the ProGen series, are commonly employed for these tasks (as illustrated in Figure \ref{protein_example1}).

In summary, LLM applications in protein research demonstrate their powerful capabilities in structure prediction, functional analysis, and sequence generation. These tools provide researchers with invaluable resources to accelerate the discovery and development of new drugs and therapeutic treatments.
\subsubsection{LLMs in Manufacturing Robotics}
The swift evolution of LLMs has significantly influenced many sectors~\cite{dai2023ad,liu2023radiology,liu2023holistic,liu2023transformation,liu2023tailoring}, yet it is in the domain of robotics where their impact is significantly pronounced. Presently, robotics stands as the foremost field that is aptly leveraging the capabilities of LLM technology.

Initially, LLMs facilitate the seamless integration of robotic responsiveness to human commands. Traditional methodologies necessitated the supplementary training of an NLP model to enable robots to autonomously interpret human directives. This approach was commonly employed in various domestic robots and smart assistants within mobile devices. Conventional NLP techniques constrained robots to execute a limited array of pre-programmed functions embedded within their processors, contingent on human input. Furthermore, these systems exhibited limitations in processing unencountered instructions during training, often resulting in inaccurate command execution. However, with the advent of LLMs, such as GPT-4, Gemini, and LLaMA, there has been a notable enhancement in robotic intelligence, providing a broader understanding and flexibility in command execution, albeit robots still exhibit certain deficiencies in fully autonomous operations.

The advent of LLMs marked a transformative phase in the domain of robotic instruction comprehension. The advanced generative capabilities of these models have endowed robots with a universal understanding of textual commands. Contemporary robotic systems are now equipped with the intrinsic ability to interpret a wide array of human directives without the necessity for further training or learning. LLMs like GPT-4 possess extensive retained knowledge across many domains and can retrieve this knowledge during inference to execute corresponding pre-programmed functions based on the received instructions.

\begin{figure*}[t]
\centering
\includegraphics[width=1\textwidth]{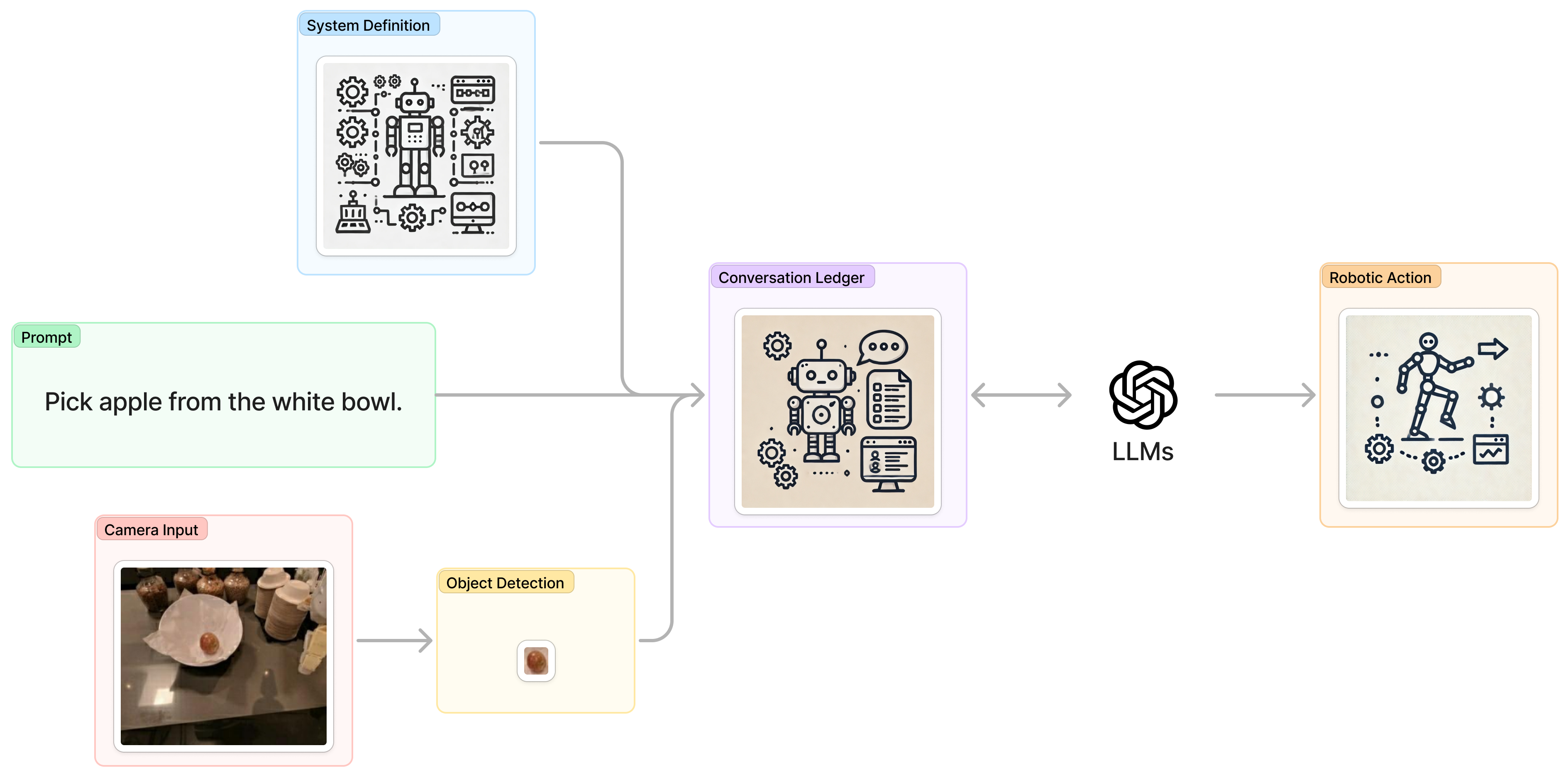}
\caption{An example of using LLMs to combine initial instructions, user commands, and vision model output from the conversation ledger to coordinate the robot's actions.} 
\label{robot_control}
\end{figure*}

A comprehensive study by Jiaqi Wang et al. \cite{wang2024largelanguagemodelsrobotics}, provides an extensive overview of the potential and challenges of integrating LLMs into robotic systems. Example can be seen in Figure \ref{robot_control}. The authors emphasize the transformative capabilities of LLMs in robotic task planning, particularly the importance of multimodal inputs such as visual and auditory data to enhance task comprehension and execution. They present a framework using multimodal GPT-4V, significantly improving robot performance in embodied tasks by combining natural language instructions with visual perceptions. This multimodal approach enables robots to understand and execute complex tasks with a higher degree of accuracy and efficiency. The study also addresses the significant challenges in deploying LLMs in robotics, including the high computational demands, energy consumption, and latency issues associated with real-time processing. To overcome these challenges, the authors suggest future research directions such as enhancing the robustness and applicability of LLM-based systems through community efforts, personalized robot control, and integrating additional modalities like hand gestures and brain signals.

Moreover, the study delves into how LLMs, through extensive pretraining on vast textual data and fine-tuning with human demonstration data, can translate high-level natural language commands into precise, actionable instructions for robots. This capability enhances the adaptability and responsiveness of robots in dynamic environments, crucial for applications in healthcare, manufacturing, and home assistance. The research emphasizes the importance of intuitive interactions and continuous learning, facilitated by LLMs, which allow robots to make real-time adjustments and improvements in task execution. This integration not only improves the accuracy and flexibility of robot control but also makes the technology more accessible and user-friendly, reducing the need for specialized programming skills.

Additionally, the autonomous reasoning capacity of LLMs significantly enhances the utility of general-purpose robots. Innovative frameworks, such as Auto-GPT~\cite{dai2023ad}, leverage LLMs to decompose overarching tasks into manageable segments, subsequently determining the optimal built-in function to address each segment sequentially, thereby accomplishing a variety of routine tasks. This methodology is exemplified by Google's RT2 robot~\cite{brohan2023rt}, which operationalizes this concept. The crux of this innovation lies in converting complex task solutions into permutations and combinations of pre-existing built-in functions. While no novel technology has been introduced at the robot hardware level, the integration of LLMs has bestowed robots with unparalleled problem-solving abilities. This advancement represents a significant breakthrough in the domain of automation, a feat that previously remained elusive despite extensive efforts. Moreover, LLMs enable robots to adjust actions based on contextual understanding, making them more adaptable and responsive to changes, which is particularly valuable in dynamic environments.

Concomitantly, an increasing number of scholars recognize the pivotal role of LLMs in advancing the field of robotics, thereby catalyzing the genesis of innovative concepts~\cite{zeng2023large, wang2024large}. Utilizing the logical reasoning and task decomposition capabilities of LLMs, robots can be trained more efficiently. Techniques such as reinforcement learning from human feedback (RLHF) and direct preference optimization (DPO) play crucial roles in this process, enhancing the model's capability to understand and align with natural language instructions. Through the large language model's reasoning faculties, intricate tasks are dissected and reconstructed using diffusion models. This process facilitates the generation of a plethora of viable and effective training datasets within virtual environments, obviating the need for gradual iterative learning in simulated or actual settings. Consequently, this enhances the robot's proficiency in learning and resolving complex scenarios. The foundational element underpinning these advancements is the integration of LLM technology. 

\begin{figure*}[t]
\centering
\includegraphics[width=0.8\textwidth]{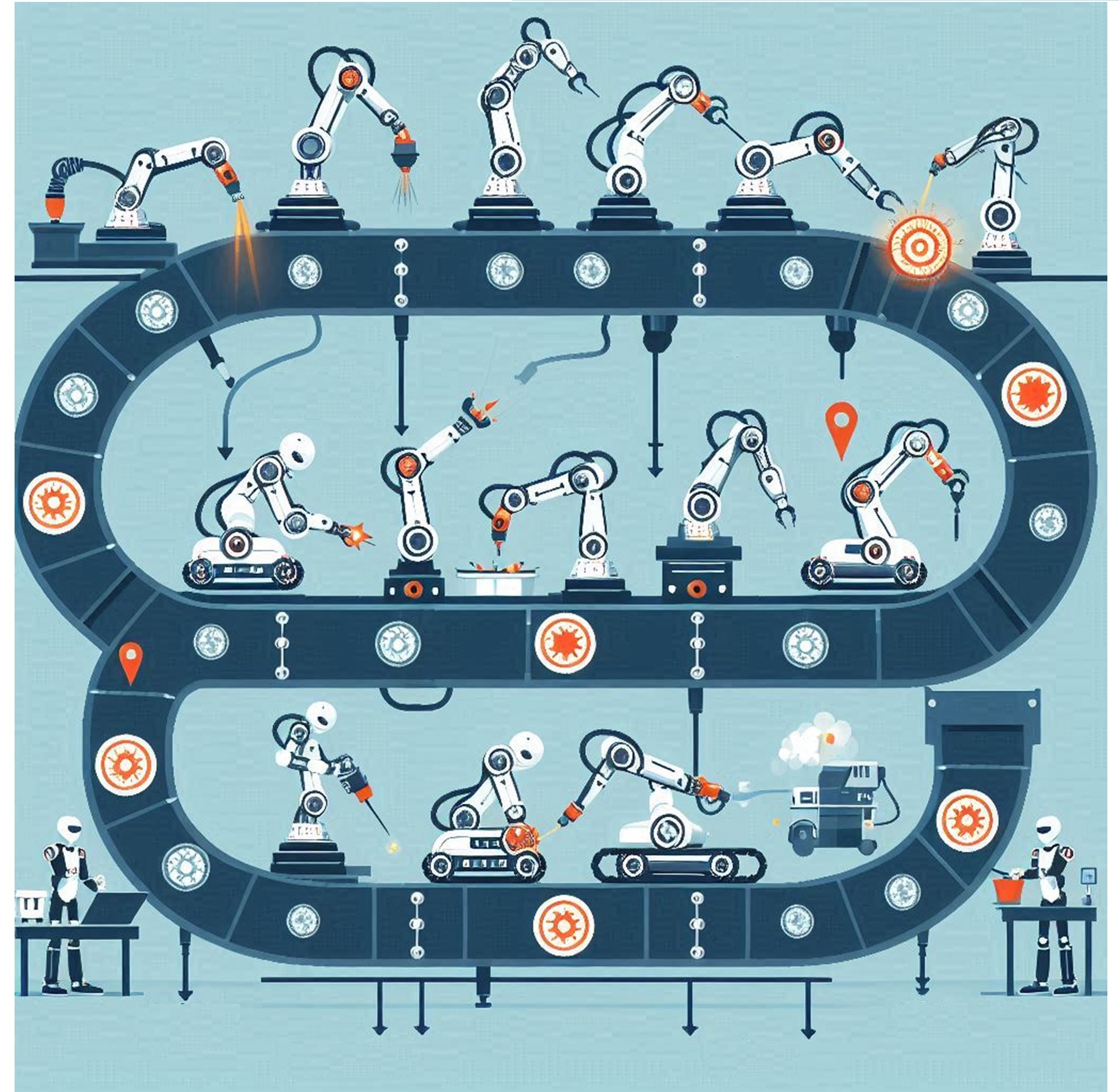}
\caption{Schematic diagram of manufacturing robots in assembly line production. Figure is generated by DALL·E 3.} 
\label{robot_manufacturing}
\end{figure*}

In the context of sim2real applications, robot training frequently confronts the challenge of insufficient annotated real-world data, necessitating reliance on simulation environments. Yet, the discrepancy between simulated and actual environments—termed the "reality gap"—often degrades the efficacy of robot training. Historically, Generative Adversarial Networks (GANs)~\cite{goodfellow2020generative, liu2023digital, liu2023dt} were employed to mitigate this issue, albeit with limitations due to their substantial data requirements and suboptimal transferability in image generation. Conversely, diffusion models~\cite{ho2020denoising} emerge as a superior alternative, with various implementation strategies aiming to enhance the controllability~\cite{zhang2023adding, li2024aldm} of image synthesis. LLMs further improve this process by enabling the generation of adaptive operation plans in real-time based on the current environment and mission requirements, thus effectively coping with uncertainties and changes in instructions.

In pursuit of this objective, numerous researchers employ LLMs to augment textual inputs~\cite{yang2024mastering}, create novel auxiliary layouts~\cite{lian2023llm}, and modify the text embedding layer~\cite{lee2024compose}, thereby facilitating more regulated image generation. Additionally, applications extend to employing text for the generation of open-world scenarios, such as robotic grasping and mobility~\cite{zhang2024hoidiffusion}, to refine robot training~\cite{zhou2024gala3d}. The growing inclination of robotics researchers towards LLMs underscores the technology's considerable promise and its evolving role in bridging the gap between simulated training and real-world application.

Furthermore, the impact of LLMs in manufacturing robotics is becoming increasingly significant as the industry seeks to capitalize on AI's potential to enhance both operational efficiency and product innovation. LLMs such as ChatGPT are not merely facilitators of communication but also showcase the capability of transforming the manufacturing workflow through advanced data interpretation and management~\cite{Rao2023}. By integrating domain-specific knowledge into LLMs and utilizing continuous learning and feedback loops, the precision and efficiency of robotics operations in manufacturing can be significantly enhanced, leading to increased productivity and reduced error rates.

\begin{figure*}[t]
\centering
\includegraphics[width=0.8\textwidth]{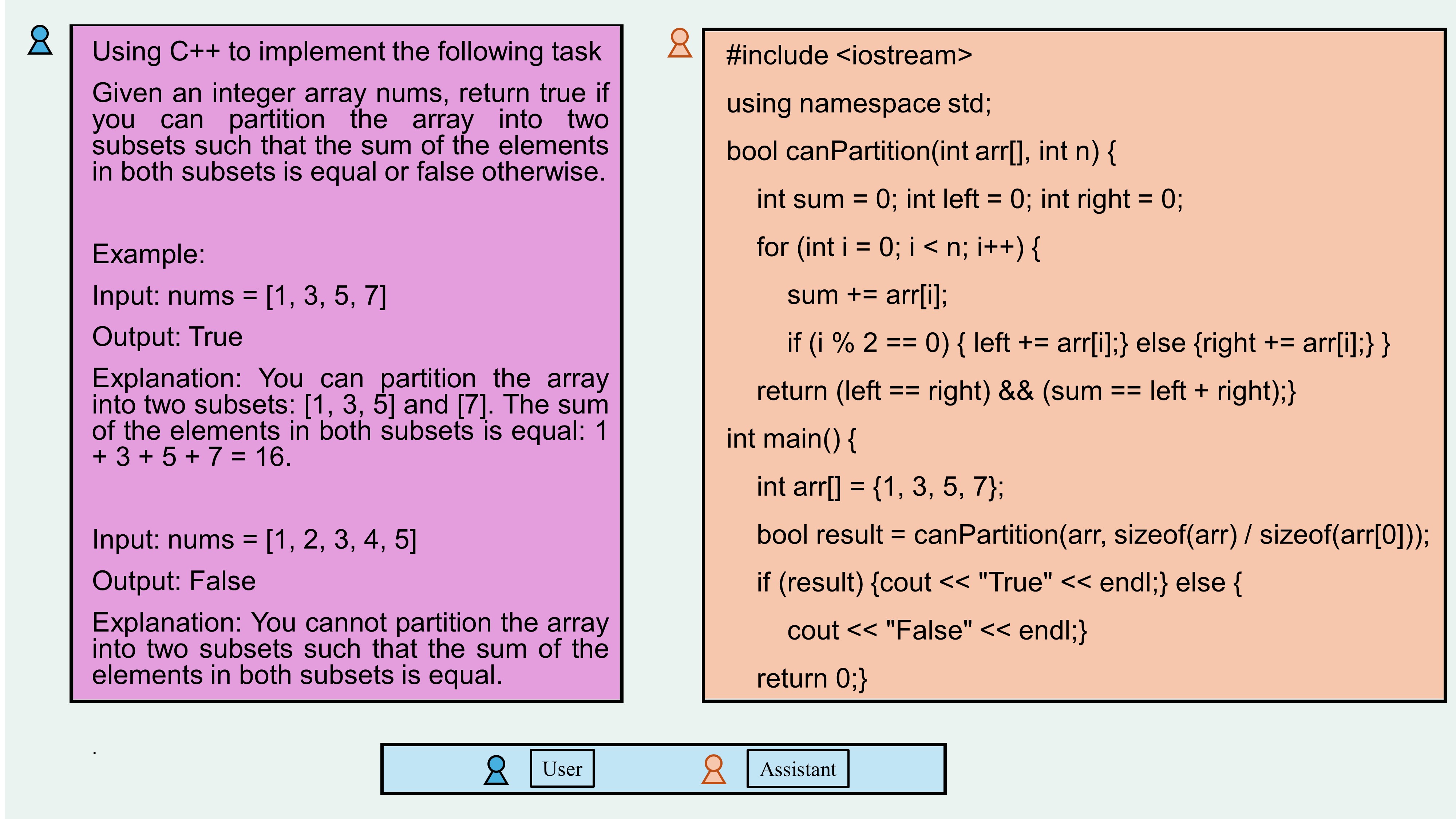}
\caption{Examples of GPT-4V generating manufacturing code~\cite{shu2024llms}. GPT4 completes the code generation very well. The generated code is not only concise and clear, without errors, but also has very good operating efficiency.} 
\label{exp_coding}
\end{figure*}
LLMs revolutionize field service operations such as predictive maintenance, where they not only predict equipment failures but also guide technicians through the maintenance process efficiently. This involves providing conversational access to technical manuals and aggregating insights from past events to recommend effective maintenance actions, significantly improving uptime and reducing costs~\cite{Steiny_Trisal_2023}. In the case of design for manufacturing, LLMs contribute significantly to computer-aided manufacturing (CAM) processes (Figure \ref{robot_manufacturing}). These models facilitate the translation of digital designs into viable manufacturing instructions, optimizing design elements to suit fabrication requirements. This capability enables engineers to efficiently explore and refine various design alternatives, ensuring optimal manufacturability and performance~\cite{makatura2023can}. LLMs dramatically improve the accessibility of complex manufacturing systems. By enabling natural language interactions with systems such as digital twins and control towers, LLMs allow even non-technical personnel to navigate intricate systems effortlessly. This reduces the learning curve and boosts productivity by making advanced technologies more accessible and responsive, which is crucial for continuous operations in the manufacturing sector~\cite{Thadikamala_2023}.

The adoption of LLMs in flexible modular production systems exemplifies their role in advancing autonomous manufacturing processes. These systems, enhanced with LLM agents, adapt dynamically to production needs and optimize processes in real time. This adaptability is essential for modern manufacturing environments that demand high efficiency and flexibility to manage complex and variable production activities~\cite{Xia_2023}. Additionally, LLMs foster a collaborative environment by integrating insights from various stages of the manufacturing process, from design to assembly line adjustments. This collaboration leads to innovative and sustainable product development, as decisions are informed by a comprehensive understanding of both design and practical manufacturing constraints~\cite{Stella_2023}~\cite{Shipps_2023}.

Besides, LLMs get involved in robot coding (Figure \ref{exp_coding}). Robot programming is one of the most fundamental but vital regions for manufacturing. It is also a field adopted by LLMs with significant application potential Where LLMs support offering advanced capabilities in code generation, bug fixing, and the automation of programming tasks. LLMs not only enhance productivity but also democratize programming knowledge, making it accessible to a wider audience with varying levels of expertise. Recently, many companies have built their LLMs as AI programmer assistants. Meta puts forward Code Llama~\cite{roziere2023code}. GitHub Copilot is developed by GitHub, OpenAI, and Microsoft for industry standards. Other programming LLMs include Codet5+~\cite{wang2023codet5+}, Tabnine, Polycoder~\cite{xu2022systematic} and etc.

One of the key areas where LLMs have made a considerable impact on robot coding is in the automation of coding tasks. Research shows that LLMs can generate code snippets and entire programs for robots with high accuracy, significantly reducing the time and effort required for designing tasks~\cite{liu2024your}. This capability is particularly beneficial in robotics engineering, where time constraints and the demand for efficiency are paramount. LLMs like CODEX~\cite{chen2021evaluating} and CodeGen~\cite{nijkamp2022codegen} perform code generation by autoregressively predicting the next token, given the previous context. This context often includes a function signature and a docstring that outlines the desired program functionality. The generated code snippet is combined with this context to form a complete function for robot executions. This approach highlights the significant capabilities of LLMs in understanding and generating code based on natural language descriptions~\cite{austin2021program,chen2021evaluating}, which evolves the design for the manipulation code of robots.
~\cite{shu2024llms} demonstrates the performance of LLMs on regular data structure and algorithms code testing as well as robot module code generation, and understanding of block diagram robot languages. They also test and attempt to generate fundamental block diagrams for robot education.

In conclusion, the integration of LLMs into manufacturing robotics represents a substantial advance toward more intelligent, efficient, and sustainable manufacturing practices. As this technology continues to evolve, it will undoubtedly unlock further opportunities for innovation and growth in the manufacturing sector.

\subsubsection{LLMs in Manufacturing Quality Control}
Quality control is a critical aspect of manufacturing that guarantees product integrity and meets customer requirements. It encompasses a range of practices and methodologies that ensure product quality is maintained throughout its lifecycle. The application of statistical techniques to oversee and regulate the manufacturing process is a fundamental basis of quality control. This universal principle has been successfully implemented across diverse manufacturing environments.

In recent years, research in the quality control field has increasingly focused on the application and advancement of deep learning technologies. Deep learning models have shown remarkable ability to handle complex, unstructured data types, such as 2D/3D camera images~\cite{Yang:DL}, thermographic images~\cite{Ho:Thermal}, acoustic signals~\cite{Cantero:ultrasonic}, and electromagnetic signals~\cite{Bhavani:electromagnetic}, significantly improving performance in detecting and classifying product defects. Additionally, the powerful feature extraction capabilities of deep learning enable the automatic identification of valuable features from large datasets, enhancing both the accuracy and efficiency of quality control processes. Stricker et al.~\cite{zhang2023large} introduced an adaptive control method based on deep reinforcement learning, which dynamically adjusts quality control strategies using real-time data, further highlighting the potential of deep learning for adaptive quality control.

Beyond improving defect detection accuracy, deep learning has also contributed to the efficiency and stability of production processes. These advancements have introduced innovative tools and techniques to the industrial sector, offering enhanced product quality and reducing production costs. By integrating advanced data-driven approaches with traditional quality control methods, deep learning is transforming manufacturing, leading to more intelligent and responsive quality assurance systems.

Conventional quality control in manufacturing has traditionally focused on ensuring that products meet predetermined standards and specifications through a series of inspections and tests performed at various stages of the production process. This approach is rooted in statistical quality control (SQC) methods, which use statistical tools and techniques to monitor and control manufacturing processes. The aim is to detect and correct problems as they occur, rather than after the fact, to prevent defective products from reaching the consumer~\cite{STUART1996203,SHAW199199}. Over time, these practices have evolved from simple inspection routines to more complex statistical analyses capable of predicting and preventing quality problems before they occur, marking a significant shift in the way quality is managed in manufacturing environments~\cite{kim2018review,MILO2015159}. As a result, conventional quality control has laid the groundwork for the development of more advanced quality control methods that incorporate real-time monitoring and automated decision-making processes~\cite{SU2019216}.

\begin{figure*}[t]
\centering
\includegraphics[width=0.8\textwidth]{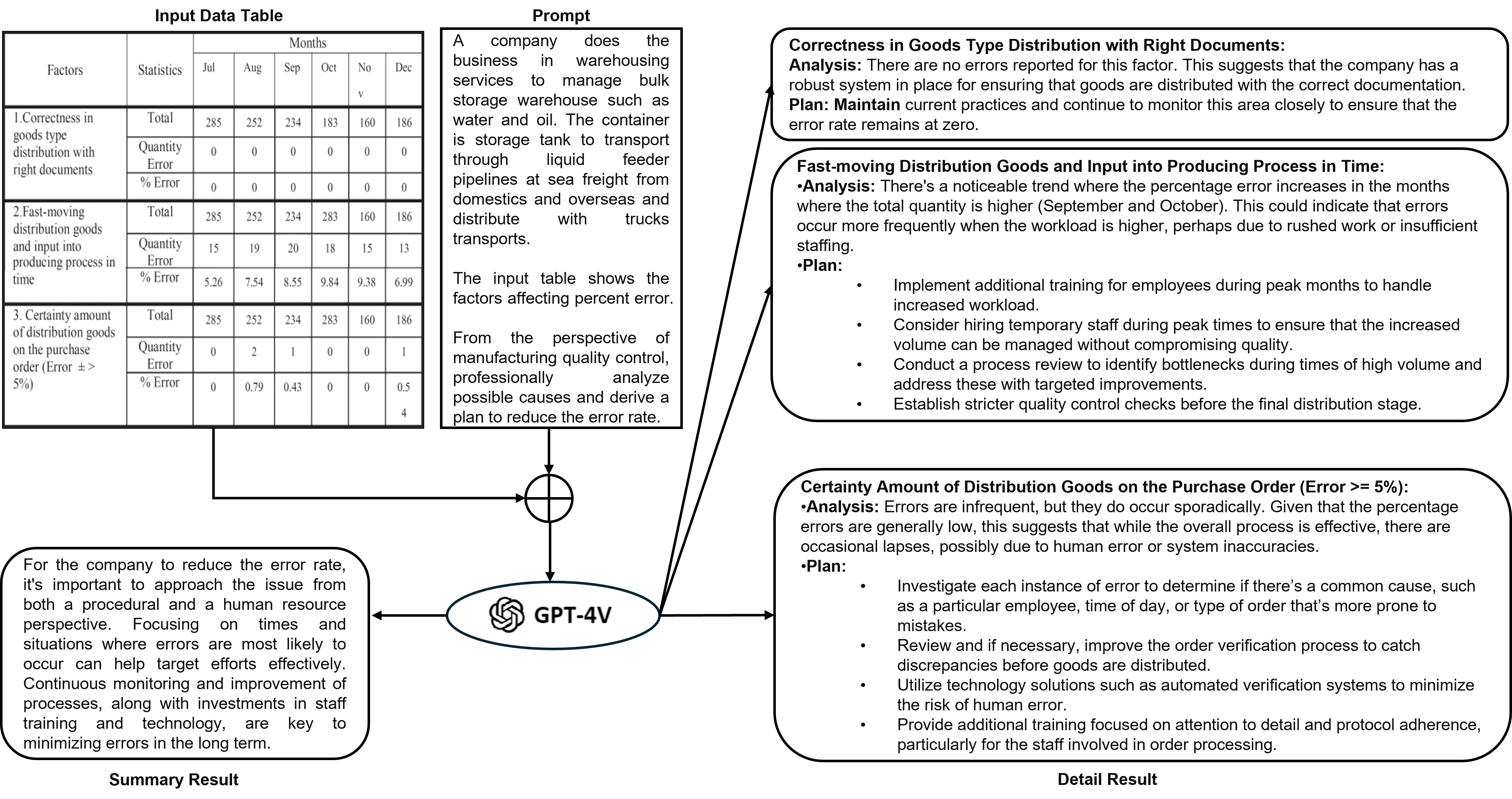}
\caption{Example for LLMs' analysis and plan on quality control. This demonstrates the ability of LLMs to undertake the planning and design of complex tasks.} 
\label{qc_example1}
\end{figure*}

\begin{figure*}[t]
\centering
\includegraphics[width=0.8\textwidth]{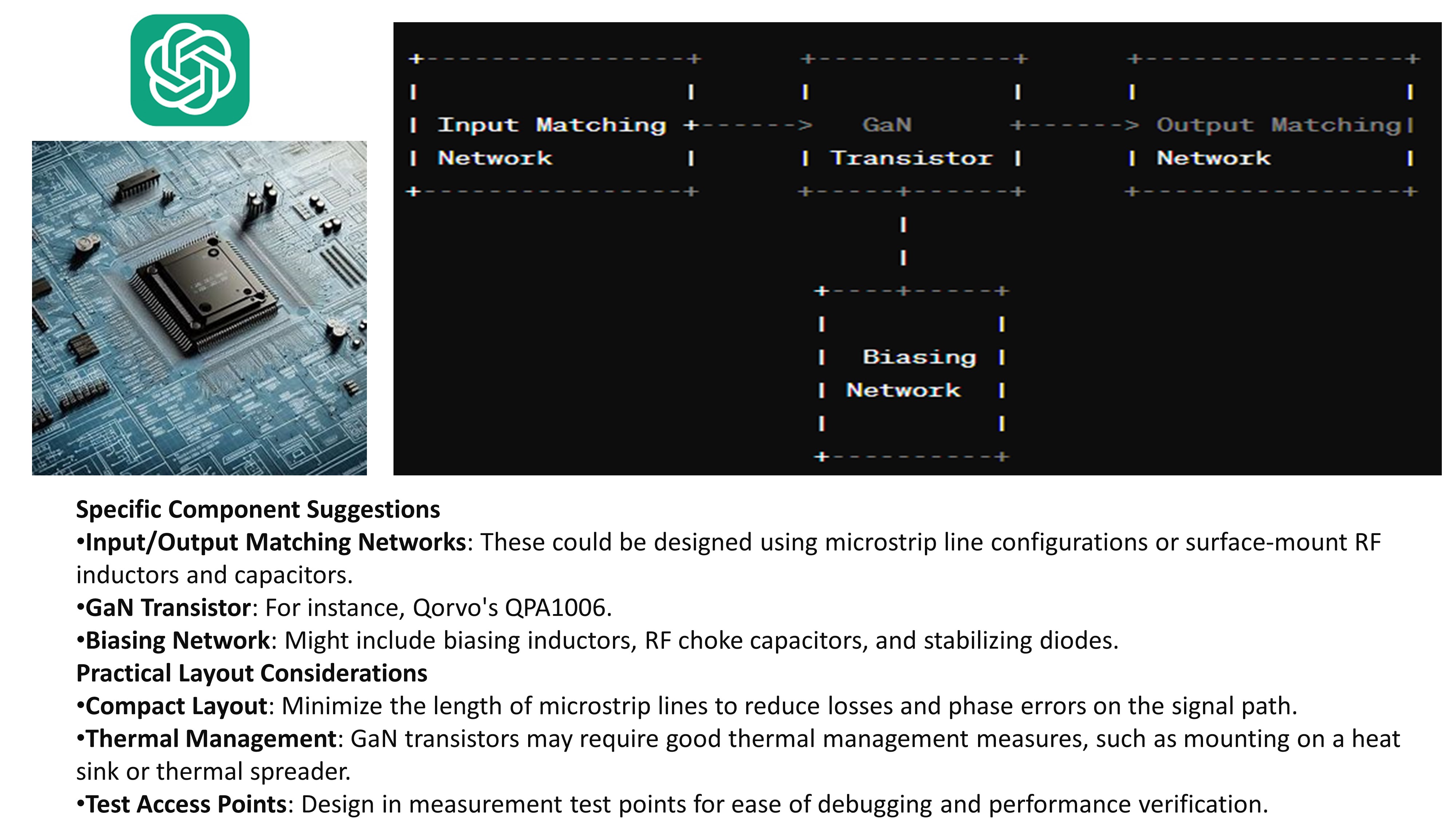}
\caption{LLMs for chip design. GPT-4V was asked to design a 3GHz-15GHz power amplifier chip and provide a specific solution.} 
\label{chip_design_example1}
\end{figure*}

The transition from conventional quality control methodologies to more technologically advanced systems is of paramount importance due to the rising expectations of consumers and the rapid development of technology. LLMs are leading this transformation by offering novel capabilities that complement and enhance traditional SQC methods, ensuring product integrity and reliability (Figure \ref{qc_example1}). In other words, the emergence of LLMs in manufacturing quality control combines traditional SQC principles with advanced AI. LLMs enhance conventional quality control practices, providing greater sophistication, efficiency, and predictive capability than previous methods. This integration signifies a new era in quality control, where statistical rigor and intelligent analysis combine to establish unparalleled levels of product quality and manufacturing excellence. 

Due to the incredible capabilities of LLMs, LLMs have the potential to revolutionize quality control processes by integrating with and building upon the fundamental principles established by conventional quality control methods. Traditional quality control methods, deeply rooted in SQC techniques, have traditionally emphasized the importance of inspection, testing and statistical analysis to maintain and improve product quality. These methods are designed to identify and mitigate defects in real time, and to ensure that final products meet pre-defined standards and specifications. LLMs can augment these SQC methods by providing a comprehensive layer of data analysis and interpretation, enabling the extraction of actionable insights from vast amounts of data, including production logs, quality inspection reports and customer feedback. Furthermore, the evolution of quality control practices from mere inspection to predictive analytics marks a significant advancement in manufacturing. LLMs are driving this evolution by enhancing the predictive capabilities of quality control systems. By analyzing historical and real-time data, LLMs can predict potential quality issues before they manifest themselves, enabling pre-emptive corrective action. This shift from a reactive to a predictive approach reduces waste, saves costs, and significantly improves the efficiency of manufacturing processes. In addition, advanced quality control methodologies emphasize the role of real-time monitoring and automated decision-making in maintaining product quality. LLMs contribute to this by facilitating informed decision-making through the analysis of complex data patterns and trends that are often beyond the human ability to recognize. Their ability to understand and generate human language enables the seamless integration of human expertise and machine intelligence, fostering a collaborative environment where decisions are both data-driven and intuitively accessible to human operators. LLMs also extend the capabilities of traditional quality control beyond the realm of statistical analysis, fostering a comprehensive quality ecosystem that encompasses every aspect of the manufacturing process. They can automate the generation and interpretation of compliance documentation, ensure regulatory compliance, and provide personalized training and support to manufacturing personnel. In this way, LLMs not only improve the technical aspects of quality control but also help to develop a quality-centric culture within manufacturing organizations.

To evaluate the capability of LLMs in the context of manufacturing quality control, we adopt the work of~\cite{7939188} and use GPT-4V to process the data. As shown in Figure \ref{qc_example1}, the input consists of two sections: the input data table and the prompt on the upper left. The input data table describes the factors that affect the percentage error of the distribution of goods from July to December. The prompt configures the background, describes the input data, and asks GPT-4V to make a professional analysis of the error rate and make appropriate suggestions. The result on the right and lower left shows a clear analysis of the possible causes and present reliable and accessible solutions to them, similar to those in the real case of the original research.

Cost control is one of the core tasks in the operation and management of manufacturing institutions. Traditional cost control methods primarily rely on human experience and judgment, which suffer from inefficiencies and subjectivity. LLMs, with their exceptional NLP abilities, bring new opportunities to cost control tasks. Research by~\cite{yu2023temporal} demonstrates that by integrating heterogeneous information such as historical stock price data, company metadata, and economic news, LLMs can accurately predict stock returns. Their study employs the GPT-4V model for zero-shot and few-shot inference, and fine-tunes the Open-LLaMA model using instructions, achieving superior performance compared to traditional ARMA-GARCH~\cite{10.5555/3052536,bahrammirzaee2010comparative,francq2004maximum,henneke2011mcmc} and gradient-boosting tree models~\cite{natekin2013gradient}. This strongly validates the advantages of LLMs in handling complex cross-sequence reasoning and multimodal information fusion tasks. Furthermore, their research introduces the concept of "Chain-of-Thoughts" by generating detailed reasoning steps, enhancing the interpretability of model outputs. With the development of LLM technology, quality control is linked to many advanced control technologies, such as IoT~\cite{petrovic2023chatgpt}, big data analytics~\cite{biswas2023focus}, and real-time monitoring systems~\cite{ajmal2024overcoming}. This illustrates a more dynamic and interconnected approach to quality control that leverages the full spectrum of digital transformation in manufacturing. This is crucial for decision-making in the manufacturing domain, where every decision requires sufficient supporting evidence.

Stock return prediction is just one example of the vast potential of LLMs in cost control. In fact, LLMs have shown great promise in various aspects of cost control. For instance, in manufacturing report analysis, researchers utilize LLMs to automatically extract key information from vast amounts of unstructured text data and generate insights for decision-making~\cite{duan2022survey, alzubi2022business}. This greatly improves the efficiency and comprehensiveness of financial analysis. Moreover, LLMs have been applied to other areas of cost control, such as budget management and resource allocation~\cite{ke2023financial}. By conducting comprehensive analyses of historical data and market trends, LLMs can provide more objective and accurate bases for cost control decisions.

Despite the remarkable progress of LLMs in cost control, existing research still has some limitations. First, current studies mainly focus on specific tasks such as stock return prediction and manufacturing report analysis, lacking systematic research on the entire process of cost control. Future research should take an end-to-end perspective, exploring the application of LLMs in various stages of cost control to build complete solutions. Second, most existing studies adopt a supervised learning paradigm, relying on large amounts of high-quality labeled data. However, obtaining such ideal data in real-world scenarios is challenging. Therefore, leveraging unsupervised or semi-supervised learning methods to maximize the potential of LLMs in cost control is an urgent problem to be addressed. Finally, the interpretability of LLMs needs to be further enhanced. Financial decisions are critical and require careful consideration. Simple model outputs are not convincing enough; clear and comprehensive reasoning processes are necessary~\cite{yu2023temporal}. This poses higher requirements for LLMs, necessitating a balance between performance and interpretability.

In conclusion, LLMs have opened up vast prospects for financial cost control, but their application is still in its infancy, with numerous challenges to be explored. Future research should focus on the following directions: first, conducting end-to-end systematic research to build comprehensive solutions covering the entire process of cost control; second, exploring unsupervised and semi-supervised learning paradigms to reduce the reliance on labeled data and improve the generalization ability of models; third, further enhancing the interpretability of models to provide credible and traceable bases for financial decisions. Only through the joint advancement of theoretical research and practical applications can LLMs truly unleash their significant value in financial cost control and the entire manufacturing domain. This requires close collaboration between academia and industry, as well as continuous exploration and innovation from all parties.

\subsection{Engineering Assistant Chatbot}
The integration of LLMs into engineering assistant chatbots represents a revolutionary advancement in Electronic Design Automation (EDA). These chatbots are designed to assist engineers by providing real-time support, guidance, and troubleshooting throughout the design process. The ability of LLMs to understand and generate human-like text allows them to interact seamlessly with engineers, offering a dynamic tool that significantly enhances productivity and accuracy on a large scale.

\subsubsection{Design Productivity}
LLMs like GPT-4~\cite{openaiIntroducingChatGPT} and Bard~\cite{manyika2023overview} are transforming engineering assistant chatbots by enabling them to understand complex engineering queries and provide precise, context-aware responses. These models, trained on vast datasets including technical manuals, design specifications, and coding standards, equip them with the knowledge required to effectively assist engineers. For example, an engineer can ask the chatbot for guidance on a specific design issue or clarification on a concept, and the LLM can provide detailed explanations or step-by-step instructions.

The large-scale impact of this technology is profound. Internal studies have shown that up to 60\% of a chip designer’s time is spent on debugging or checklist-related tasks, such as verifying design specifications, constructing testbenches, defining architectures, and managing tools or infrastructure~\cite{10.5555/3361338.3361354}. By integrating LLMs into chatbots, engineers can receive immediate assistance, reducing the time spent waiting for expert advice. This can significantly improve overall productivity without impacting others, allowing engineers to focus more on brainstorming, designing hardware, and writing code.

\subsubsection{Real-Time Problem Solving and Knowledge Sharing}
One of the critical contributions of LLMs in engineering assistant chatbots is their ability to provide real-time problem-solving assistance. When an engineer encounters an error or a bug in their design, the chatbot can quickly analyze the issue and offer potential solutions. This immediate feedback loop significantly reduces downtime and accelerates the debugging process. By leveraging extensive training data, the LLM can recognize patterns and suggest fixes that might not be immediately apparent to a human engineer.

Furthermore, these chatbots serve as a repository of corporate knowledge, extracting information from internal design documents, codebases, recorded design data, and technical communications. This centralization of knowledge ensures that engineers have access to the most relevant and up-to-date information, fostering a more efficient and informed design process. It also facilitates knowledge sharing across global teams, ensuring that expertise is readily available regardless of geographical location.

\subsubsection{Enhanced Collaboration and Integration with Tools}
LLMs enhance the functionality of engineering assistant chatbots by enabling seamless integration with existing EDA tools and systems. Through APIs and custom connectors, these chatbots can access design files, run simulations, and even control aspects of the design software. For instance, an engineer can instruct the chatbot to modify a component in the design, and the chatbot can execute the changes directly within the EDA tool. This integration reduces the need for manual interventions, minimizing errors and saving time.

Moreover, LLM-powered chatbots help coordinate collaborative efforts among team members by logging interactions and decisions made during the design process. This ensures that all team members are updated and on the same page, enhancing teamwork and ensuring effective knowledge sharing. By acting as a centralized communication hub, these chatbots facilitate smoother project management and collaboration.

Practical applications of LLMs in engineering assistant chatbots highlight their transformative impact. For instance, NVIDIA’s ChipNeMo~\cite{liu2023chipnemo} utilizes LLMs to provide domain-specific support in chip design. Engineers using ChipNeMo receive real-time guidance and automated support tailored to their specific needs, resulting in more efficient design cycles and reduced time to market for new products.

Similarly, RapidSilicon’s upcoming tools leverage LLMs to enhance engineering assistants in hardware design. These tools would provide detailed technical assistance, automate routine tasks, and facilitate complex design processes, leading to a more streamlined and productive engineering workflow. The ability of LLMs to quickly adapt and provide relevant information makes them invaluable in maintaining high-quality standards and improving overall efficiency.

\subsubsection{EDA Script Generation}
Recently, the integration of AI with EDA algorithms has begun to revolutionize productivity in chip design. Even though traditional EDA tools have been advanced with the capacity to provide industrial solutions, they still leave many language-intensive tasks, such as coding and error classification, reliant on manual efforts. This gap is where LLMs are making a transformative impact. 

The advent of both commercial (e.g., GPT family~\cite{openaiIntroducingChatGPT}, Bard~\cite{manyika2023overview}) and open-source (e.g., Vicuna~\cite{chiang2023vicuna}, Llama family~\cite{touvron2023llama,touvron2023llama2,roziere2023code}) foundational LLMs opens new avenues for automating these language-intensive tasks. These models excel in code generation, engineering problem-solving, analysis report generation, and error classification, significantly boosting design efficiency and precision. Moreover, domain-specific LLMs like BloombergGPT~\cite{wu2023bloomberggpt} and BioMedLM~\cite{bolton2024biomedlm} have shown superior performance in their corresponding specialized tasks, while open-source models offer promising capabilities in hardware design, outperforming general-purpose models in certain areas. Tailoring LLMs to specific domains helps mitigate security risks involved in sharing proprietary chip design data with external parties. This is well addressed by ChipNeMo~\cite{liu2023chipnemo}, developed by NVIDIA. ChipNeMo leverages domain-adapted LLMs to automate and enhance various facets of chip design. By employing domain-adaptive techniques like continued pre-training and model alignment specifically tailored for chip design, ChipNeMo excels in EDA script generation. For instance, in direct applications, ChipNeMo’s domain-adapted models have been shown to outperform general-purpose models like GPT-4V in generating EDA scripts, demonstrating the immense potential and capability of LLMs in this high-tech field. In addition, a study by Jason et al.~\cite{10299874} from New York University explores the integration of LLMs in hardware design, focusing on the end-to-end translation of specifications into Hardware Description Languages (HDLs) like Verilog. This process, typically time-consuming and prone to errors, is now being automated using conversational LLMs such as ChatGPT. The study involved co-designing an 8-bit accumulator-based microprocessor, where the LLMs assisted in various stages of the design, including instruction set architecture (ISA) definition, component implementation, and bug fixing. The result was a processor that was sent to tapeout, marking a significant milestone as the first wholly-AI-written HDL for tapeout.

These advancements underscore the critical role of LLMs in enhancing chip design processes. By reducing the reliance on external LLM services, which pose security risks for proprietary data, and utilizing in-house adapted models, companies like NVIDIA are not only safeguarding their data but are also setting new benchmarks in the chip design industry. The successful implementation of LLMs in chip design, as demonstrated by ChipNeMo and , represents a significant forward leap in how semiconductor companies can leverage AI to optimize their design processes, embodying the future of chip manufacturing technology. 

However, the application of LLMs in chip design and manufacturing also faces challenges and limitations. Due to the technical complexity and engineering requirements, LLMs must possess extensive domain knowledge to be effectively deployed. Moreover, the design solutions proposed by LLMs may be limited by pre-defined specifications and constraints, potentially leading to less flexible or suboptimal designs. Thus, blending the intelligence of LLMs with human expertise remains a critical challenge.

A a sectional summary, the integration of LLMs in chip design and manufacturing has catalyzed substantial enhancements and unveiled significant potential for the industry. These models, by automating complex, language-intensive tasks such as generating code, EDA scripts and classifying errors, enrich the design process with advanced algorithms and NLP capabilities. This not only elevates the efficiency of the chip design workflow but also facilitates the broader dissemination and refinement of programming expertise. Looking ahead, the functionality and relevance of LLMs are poised to improve through targeted strategies like domain-adaptive pre-training and precise model alignment. The exploration of innovative applications, including engineering assistant chatbots, specialized scripting tools, and comprehensive defect analysis, promises to drive the sector toward heightened automation and refinement. Consequently, the future of LLMs in chip design and manufacturing appears exceptionally promising, likely to significantly influence the evolution and breakthroughs within the realm of electronic technology.

\begin{figure*}[t]
\centering
\includegraphics[width=1.0\textwidth]{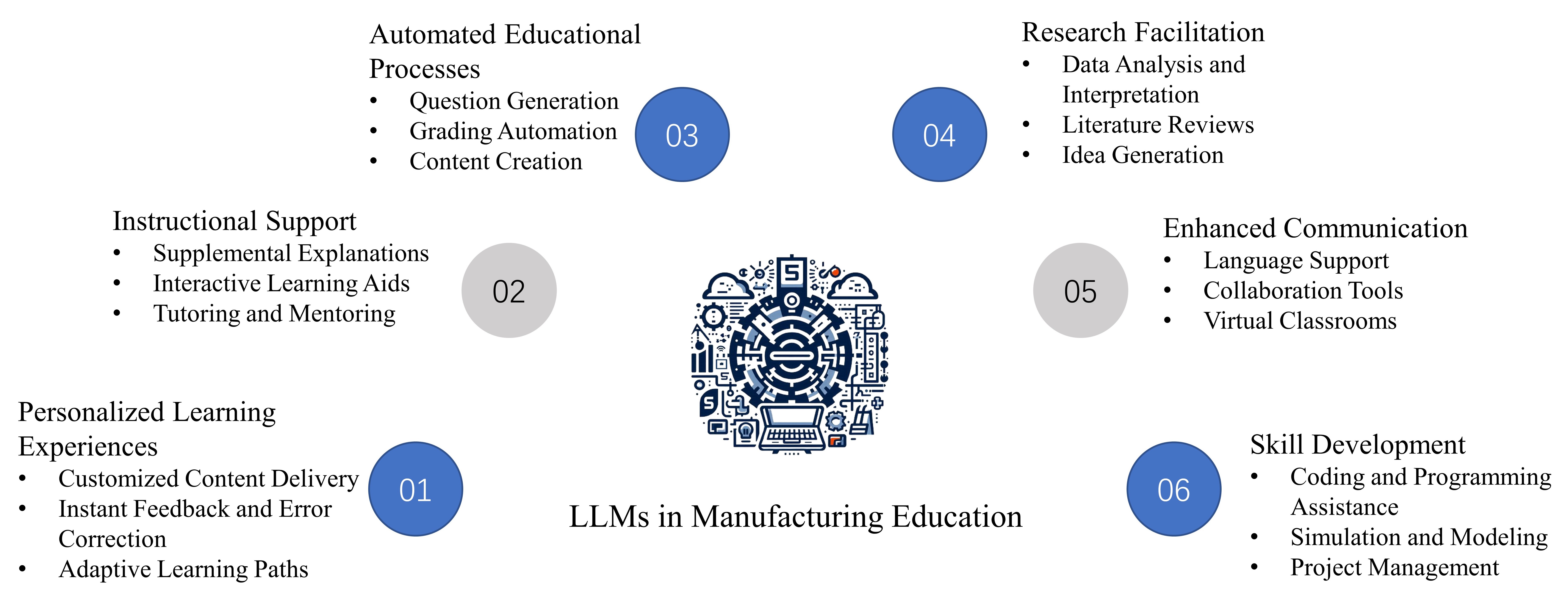}
\caption{Examples of some applications for LLMs in manufacturing education. They can also extend to general education besides manufacturing region. The logo in this figure is generated by Image Creator from Microsoft Designer.} 
\label{exp_edu}
\end{figure*}

\section{LLMs in Manufacturing Supply Chain Management}
In the realm of supply chain management, leveraging LLMs has shown significant promise in enhancing various facets of manufacturing operations. This advanced technology aids in refining processes ranging from demand forecasting to production optimization and logistics management, significantly enhancing operational efficiency and responsiveness. Figure 20 is a detailed logistic workflow empowered with LLMs.

\subsection{Demand Forecasting}

By analyzing historical sales data, market trends, and seasonal factors, LLMs enable manufacturing firms to accurately forecast future product demands. This precision in analysis helps companies refine their production schedules to better align with anticipated market needs.
Furthermore, LLMs excel in processing customer and market feedback, providing manufacturers with deep insights into consumer demand fluctuations~\cite{loukili2023sentiment}. It enables sentiment analysis to quantify textual data, capturing market sentiments effectively. By analyzing consumer feedback through the assessment of positive, negative, or neutral tones in text, LLMs facilitate a nuanced understanding of market demands. This quantitative approach allows manufacturers to model consumer preferences more accurately, aiding in the alignment of product development and marketing strategies with current consumer trends, thereby enhancing business responsiveness to market dynamics~\cite{zhao2024revolutionizing}. Also, LLMs greatly enhance the supplier evaluation process by analyzing suppliers' credit histories, delivery records, and quality control documents. Before the LLMs, the practitioner implemented diverse policies across suppliers, which may not align with the linear models typically assumed in their frameworks. For example, Moutaz's study analyzes four underexplored supplier credit policies, differentiated by fixed credit terms and those varying with order volume. It covers scenarios of full or partial upfront payments, enabled by digital inter-organizational connections, providing closed-form solutions for all policies~\cite{khouja1996optimal}. The scale of LLMs and quality of data representations support businesses in selecting the most reliable suppliers, thereby enhancing the reliability and efficiency of the supply chain.

\subsection{Production Optimization}

LLMs analyze data throughout the manufacturing process to identify inefficiencies and propose improvements, optimizing the performance of production lines. Jeong has examined that, in a complex manufacturing environment with multiple production lines and machines of varying capacities, tasks demanding specific materials, machinery, and precise delivery schedules are continuously received. An LLM trained on historical scheduling data can propose optimal schedules for new tasks (Figure \ref{logic_example1} and \ref{logistic_solve}), taking into account both task specifics and the current state of production lines. This approach highlights the potential of AI and data-driven solutions in addressing complex manufacturing challenges~\cite{jeong2023digitalization}. The factory production setting inherently generates substantial amounts of data, characterized by highly reliable labels due to stringent quality control processes that correct any discrepancies. This reliability allows for the use of local data to customize and train localized LLMs specifically for Production Optimization. By leveraging this localized LLM, manufacturing operations can enhance efficiency and output through data-driven decisions, reflecting a practical application of AI in industrial contexts.

\subsection{Fault Prediction and Maintenance}
In predictive maintenance, LLMs analyze operational data from equipment in real time to forecast potential equipment failures, facilitating timely maintenance actions that reduce machine downtime and enhance overall productivity. The system incorporates several key components (Figure \ref{fault_pred}):

Validation Scripts: These automated scripts are crucial immediately after task completion. They check the accuracy and adherence of the output to predefined standards, ensuring that each task meets the set expectations and maintains quality assurance.
AI Agent: This component serves as the operational core, interpreting task directives and executing them accordingly. It continuously learns from the outcomes and feedback provided by the validation scripts, adapting its methods to improve future performance and accuracy.
Logging: This process involves the systematic recording of all operations and their outcomes. Logging is essential for tracking the AI agent’s decisions and results, which aids in analyzing trends, diagnosing issues, and refining the system for better future performance~\cite{badini2023assessing,cao4741492managing}.
The processes described above constitute the core capabilities enabling automated error correction within factory operations, significantly enhancing efficiency. By minimizing human subjectivity, these systems reduce the likelihood of misjudgments. Furthermore, in instances where machines encounter issues beyond their processing capabilities, the comprehensive logging of operations facilitates detailed troubleshooting. The integration of LLMs into these processes markedly improves operational efficiency, showcasing the transformative potential of AI in industrial settings.

\begin{figure*}[t]
\centering
\includegraphics[width=0.8\textwidth]{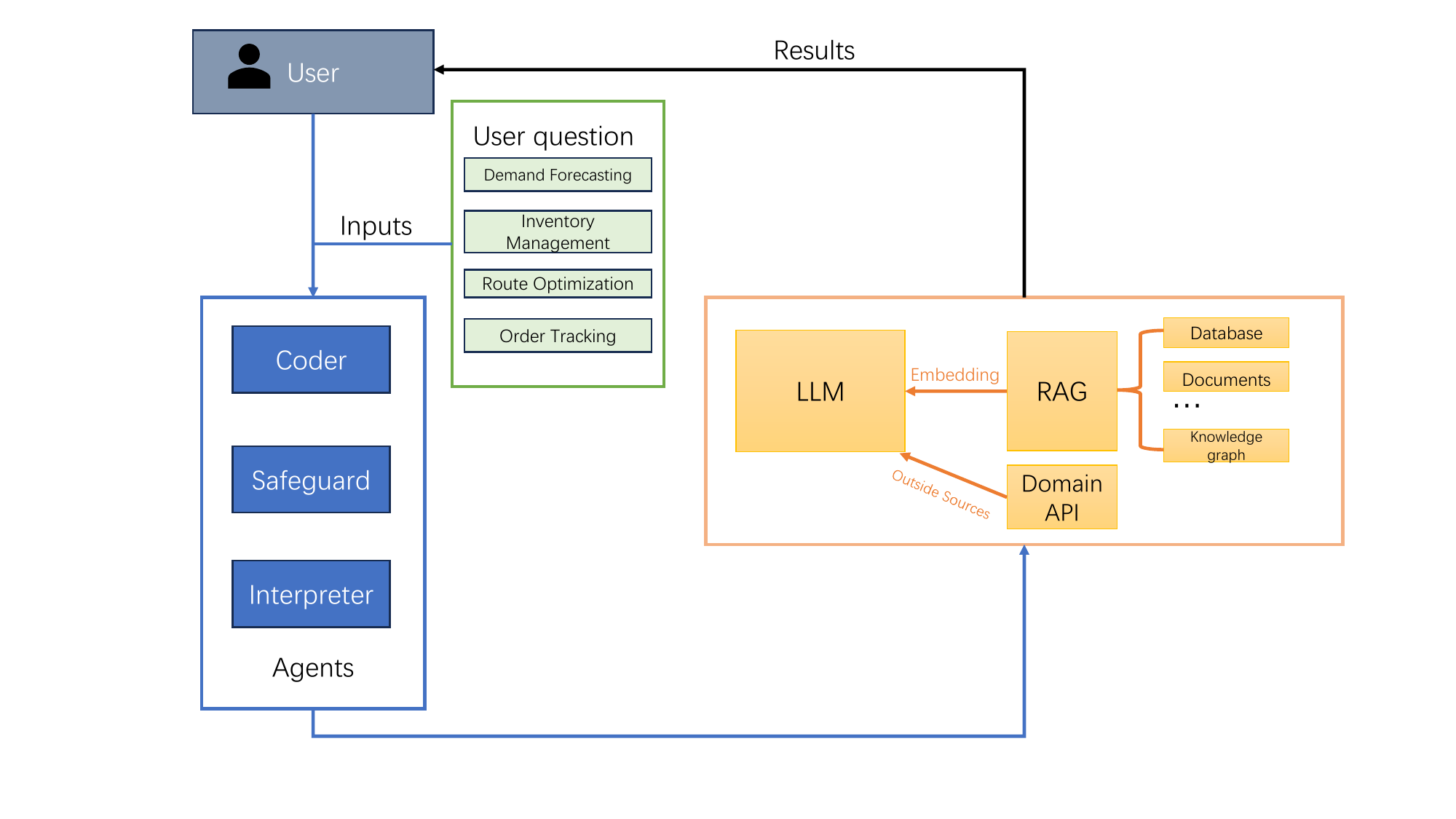}
\caption{Logistic workflow empowered with LLMs.} 
\label{fault_pred1}
\end{figure*}

\begin{figure*}[t]
\centering
\includegraphics[width=0.8\textwidth]{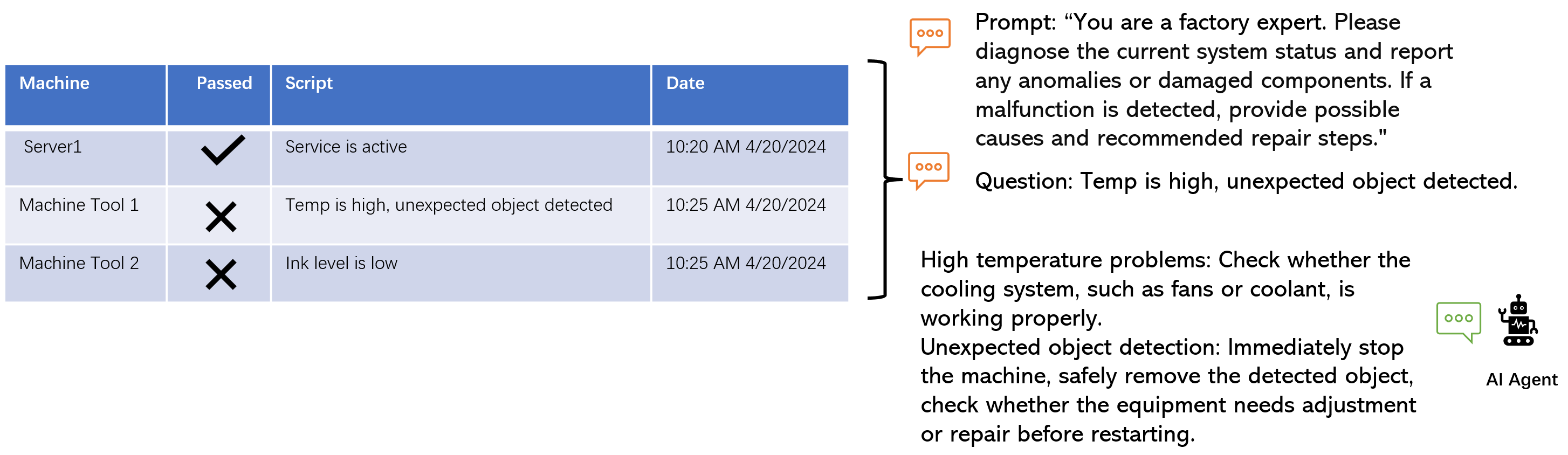}
\caption{Example for LLMs' analyze operational data from equipment in real-time to forecast
potential equipment failure.} 
\label{fault_pred}
\end{figure*}

\subsection{Logistics and Distribution Optimization}

LLMs serve not only as the cognitive hub of factory operations but also possess the capability to interface with external APIs to access real-time information. This integration enables the resolution of real-time issues, leveraging timely data to enhance operational decision-making and responsiveness in a dynamic industrial environment. By integrating external APIs for weather conditions and traffic flow, LLMs optimize delivery routes and schedules, significantly reducing transportation costs and improving delivery efficiency~\cite{zhang2023ecoassistant,Tesfagiorgis1801354}.

\begin{figure*}[t]
\centering
\includegraphics[width=0.8\textwidth]{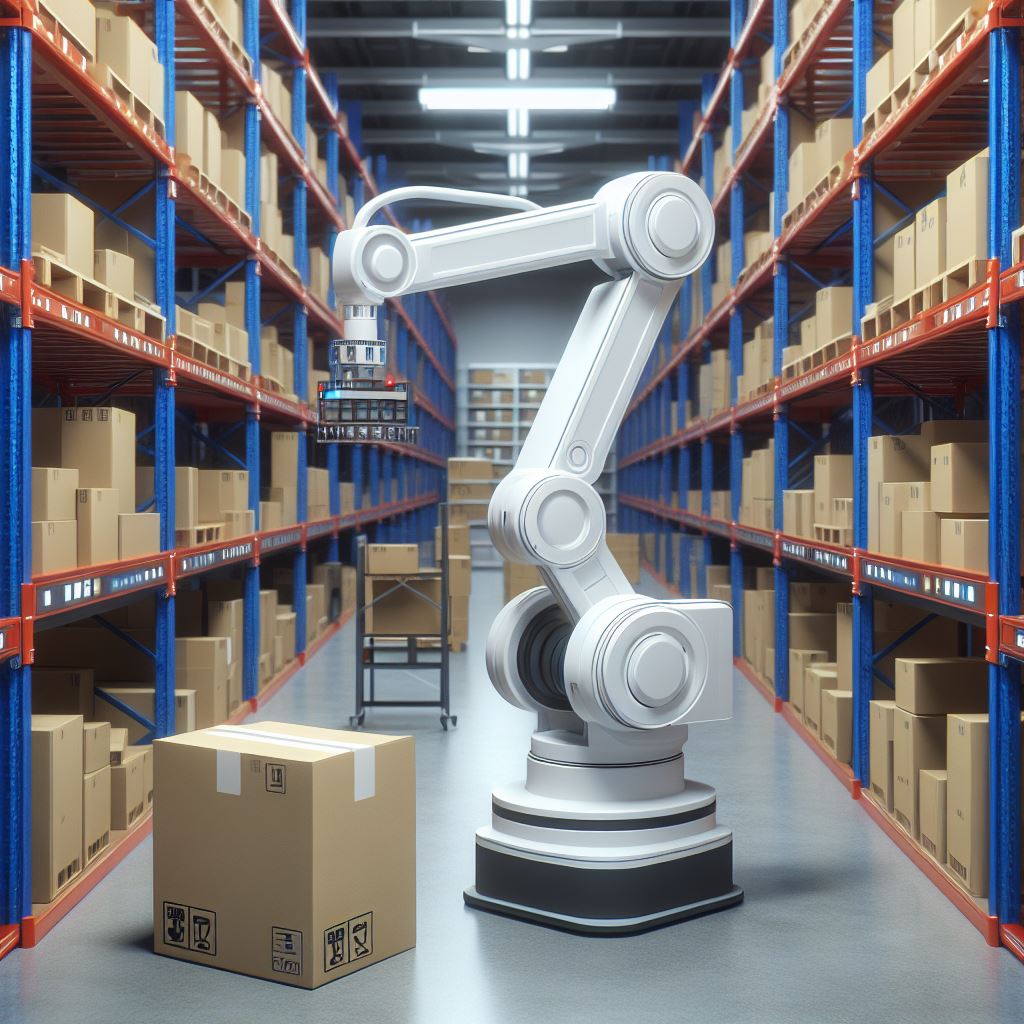}
\caption{Example for LLMs empowered Logistics within the factory. Figures are generated by DALL·E 3.} 
\label{powerarm}
\end{figure*}
Furthermore, LLMs not only exhibit high accuracy but also demonstrate proficiency in zero-shot learning, enabling them to make informed inferences in response to rare or unprecedented situations. This attribute renders LLMs particularly advantageous for the logistics industry, which demands constant operational readiness and adaptability. Additionally, the implementation of LLMs in such settings is cost-effective, as their capacity to automate intricate decision-making processes significantly reduces the reliance on manual oversight, thereby decreasing associated operational expenses.
The applications of LLMs in manufacturing, as evidenced by the research conducted by Badini et al.in the field of additive manufacturing, demonstrate their capacity to handle technical queries effectively, especially in troubleshooting and optimizing Fused Filament Fabrication (FFF) processes using thermoplastic polyurethane. The research highlighted the LLMs' high precision, accuracy, and organizational skills in managing common printing issues such as bed detachment, warping, and stringing~\cite{hadi2023survey}
\begin{figure*}[t]
\centering
\includegraphics[width=0.8\textwidth]{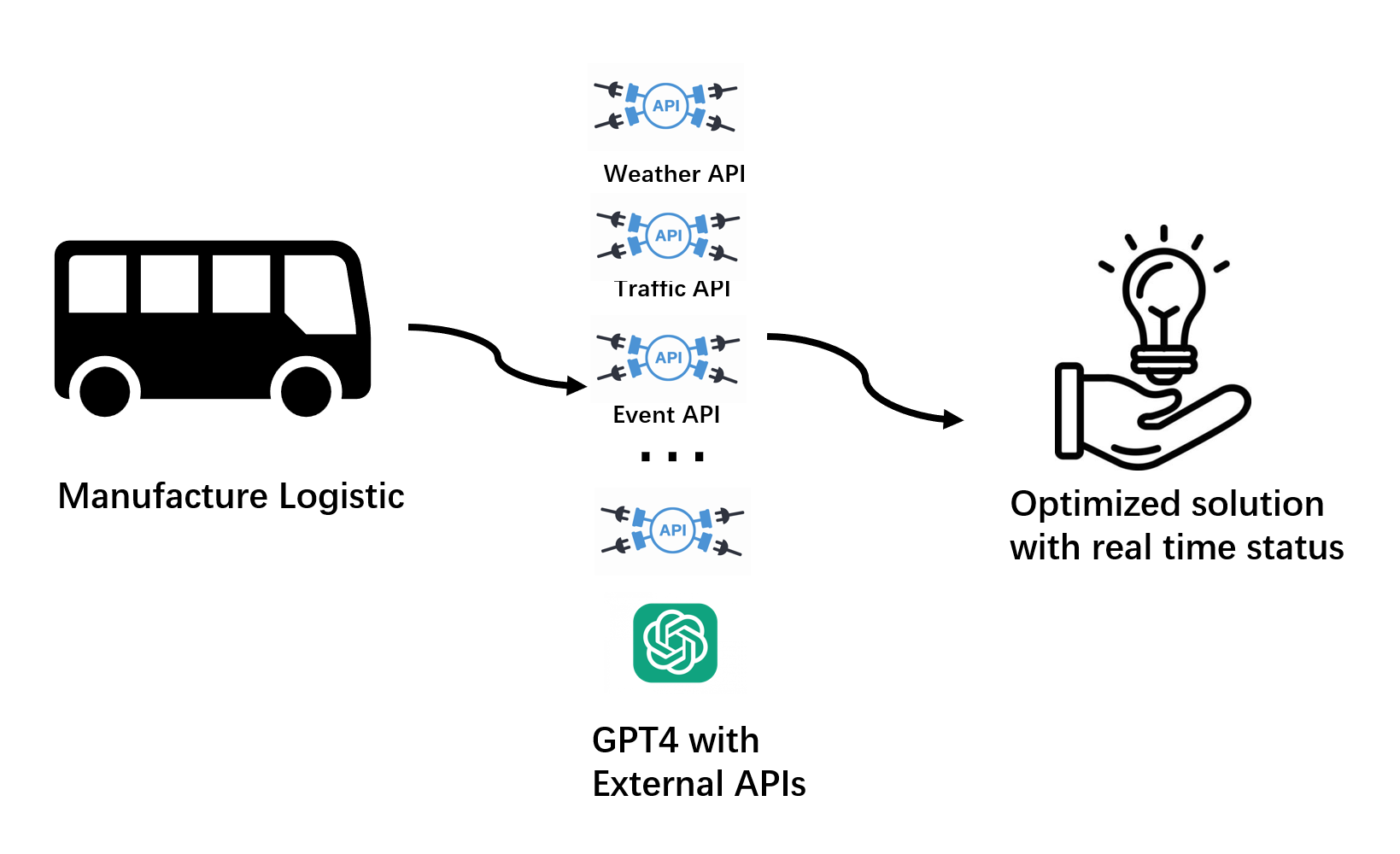}
\caption{An example of LLMs using external APIs and GPT to solve real-time logistics problems.} 
\label{logistic_solve}
\end{figure*}

The integration of LLMs into manufacturing processes, particularly additive manufacturing, can greatly enhance operational efficiency and quality (Figure \ref{powerarm} and Figure \ref{whole_process}). The real-time recommendations provided by LLMs can be instrumental in making manufacturing operations more responsive and adaptive to dynamic market conditions and technological advancements.

\section{LLMs in Manufacturing Education}
\subsection{LLMs Impact on Future Education}
The integration of LLMs into the educational field, specifically within engineering (including manufacturing) practices, marks a significant progression in instructional strategies and learning experiences. Emerging research provides an all-around view of how these advanced AI technologies are reshaping the landscape of engineering education. Figure~\ref{exp_edu} shows the existing and potential application and aspects to combine LLMs in manufacturing education. They can be summarized as offering personalized learning, instructional support, innovative research methodologies, Automated Educational Processes, Enhanced Communication, Skill Development and etc.

Studies have emphasized the potential of LLMs to offer highly personalized learning experiences and instructional support. For instance, LLMs are particularly effective in correcting errors and providing instant feedback, which is especially beneficial for students in the early stages of learning. These models excel at addressing common beginner mistakes and help students grasp complex engineering concepts by offering tailored explanations and hints, rather than direct answers, thereby encouraging independent problem-solving~\cite{abedi2023beyond}. 

With powerful plugins like Wolfram Alpha for solving mathematical problems and interpreting code, LLMs expand their functionality, transforming into comprehensive educational tools~\cite{abedi2023beyond}. These models can adapt content delivery to suit individual learning styles and paces~\cite{wang2024largelanguage}. For example, ~\cite{kuo2023leveraging} demonstrated how LLMs can generate adaptive learning paths based on students' knowledge assessments, while ~\cite{kabir2023llm} shows how LLMs automatically generate questions for subsequent Learning Objects (LOs) when mastery of a specific topic is demonstrated. LLMs can also collect multidimensional data to help educators assess students' learning progress and adjust teaching content accordingly, offering personalized growth services~\cite{xu2024large}.

\begin{figure*}[t]
\centering
\includegraphics[width=0.8\textwidth]{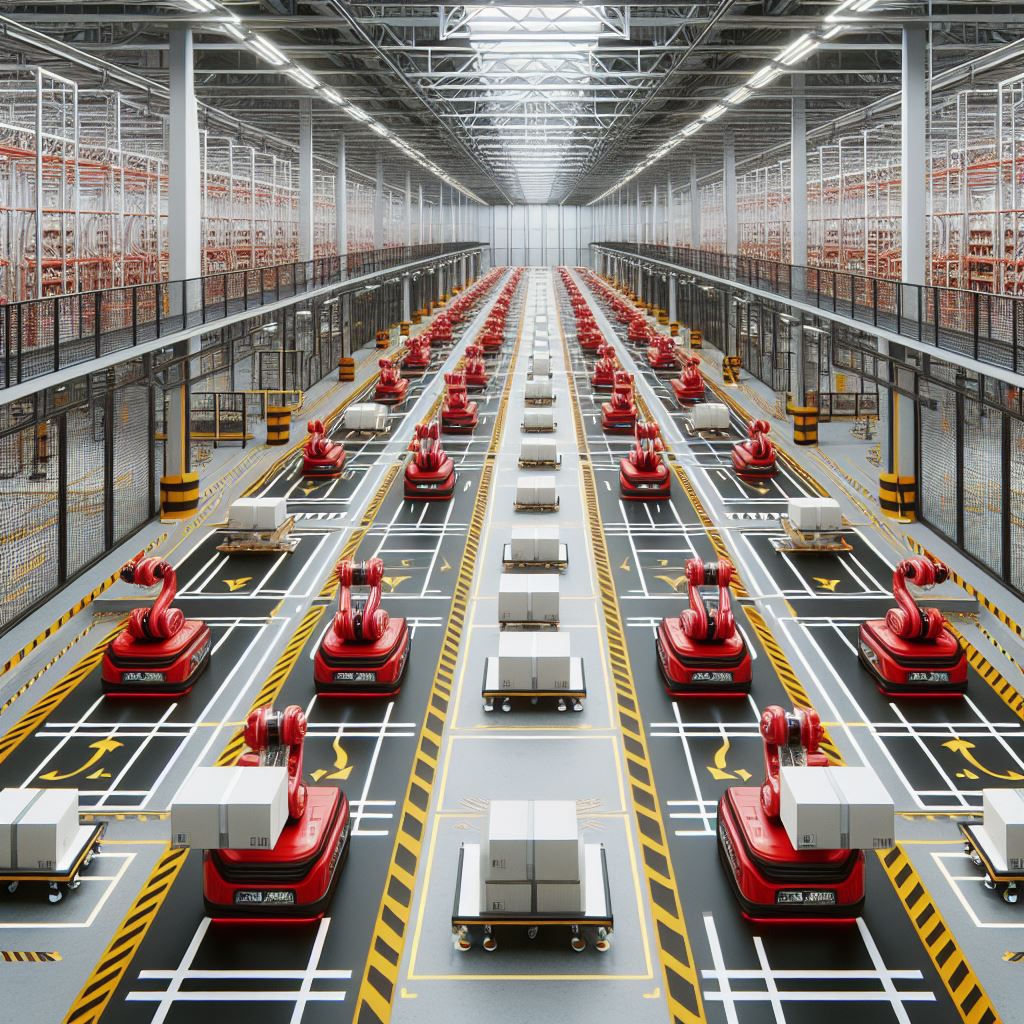}
\caption{Example of LLMs empowered logistics within the factory. Figures are generated by DALL·E 3.} 
\label{logic_example1}
\end{figure*}

In the field of mechanical engineering, research by Tian et al.~\cite{tian2024assessing} highlights the effectiveness of LLMs in addressing conceptual questions across various mechanics courses. Their study shows that GPT-4V, in particular, outperforms earlier models and even human counterparts in several areas, although it still faces challenges with symbolic calculations and tensor analysis. This underscores the growing potential of LLMs as valuable tools in engineering education and research, while also pointing to the need for further advancements to fully realize their capabilities. For example, many students turn to LLMs for supplemental explanations. Models like GPT-4 provide detailed explanations and hints for difficult concepts, which would otherwise take a considerable amount of time to find in textbooks or search results. LLMs provide precise explanations that save time and promote independent problem-solving. 

Moreover, LLMs are not limited to predefined instructions; they can interact and communicate with humans and other AI agents~\cite{park2023generative}. These qualities make LLMs ideal for use in interactive learning environments, where they can create simulations and visualizations to deepen students' understanding of engineering principles. Simulations powered by LLMs are used across various fields~\cite{park2022social,li2023you,kovavc2023socialai,jinxin2023cgmi}, demonstrating their potential as a new paradigm for simulation-based education by endowing agents with human-level intelligence~\cite{gao2023large}. These LLM-based systems excel in educational tasks such as self-learning tools and virtual tutors~\cite{garcia2024review}, assisting with automatic answer grading~\cite{ahmed2022application}, generating explanations, creating questions, and resolving problems. They also help synthesize information and enhance students' abstraction skills, particularly when summarizing texts~\cite{phillips2022exploring}.

Recently, researches have invented innovative research methodologies enpowered by LLMs. LLMs are used to support data analysis. Traditional approaches depend on human expertise. However, LLMs allow a paradigm shift towards more automated, intelligent systems. These models have the potential to analyze vast amounts of qualitative data, such as user feedback, software documentation, and development logs, to uncover valuable insights. By enhancing data analysis efficiency, LLMs also add depth and nuance to the interpretation of qualitative data, achieving a level of understanding that was previously difficult to attain~\cite{rasheed2024can}. There have been several researches exploring the potential of LLMs in qualitative data analysis~\cite{chew2023llm,dai2023llm,xiao2023supporting}.
Apart from data analysis, LLMs have been used for literature review. Literature reviews constitute an indispensable component of research endeavors. However, they often prove laborious and time-intensive. Models such as GPT-4 can help compile and summarize relevant academic literature for research projects while saving magnificent amount of time. Findings in ~\cite{zimmermann2024leveraging} indicate that ChatGPT aids researchers in swiftly perusing vast and heterogeneous collections of scientific publications, enabling them to extract pertinent information related to their research topic with an overall accuracy of 70\%.

\begin{figure*}[t]
\centering
\includegraphics[width=1.0\textwidth]{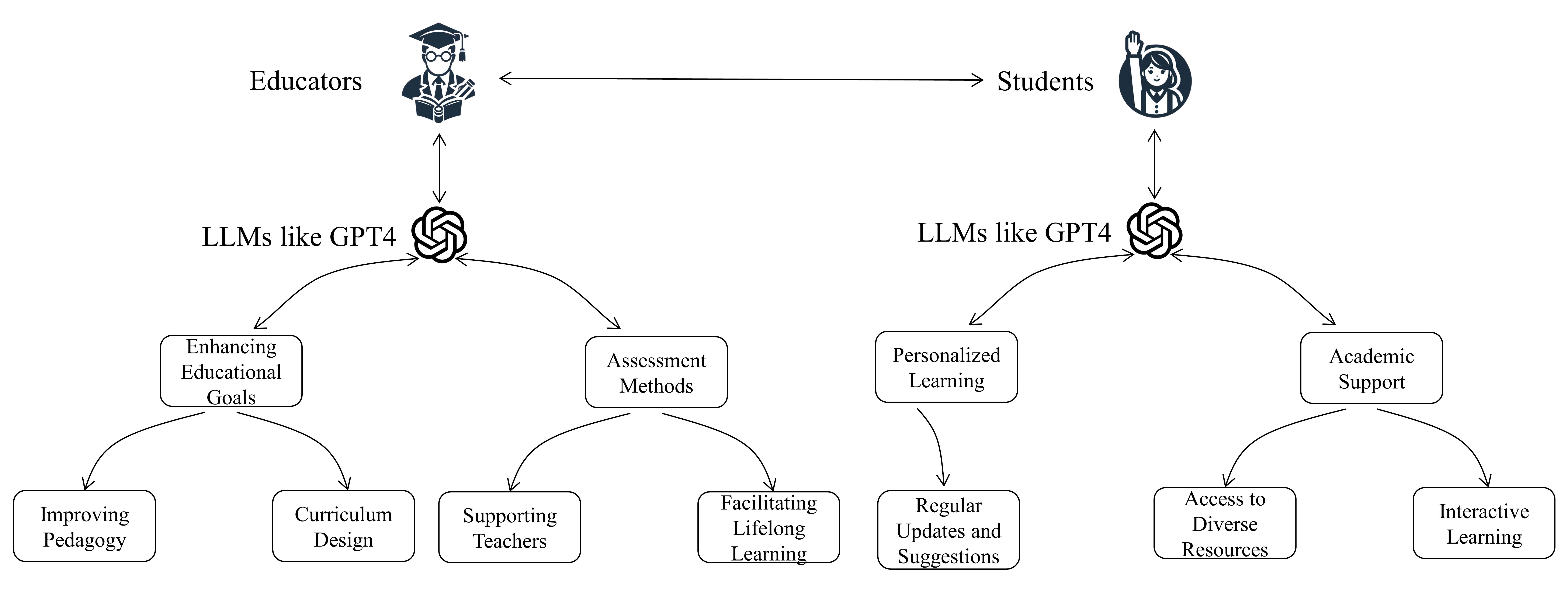}
\caption{Examples of interaction between LLMs and educators as well as LLMs with students. By assisting with LLMs, future education will be revolutionized for better quality and experience.} 
\label{exp_edu2}
\end{figure*}

\begin{figure*}[t]
\centering
\includegraphics[width=0.8\textwidth]{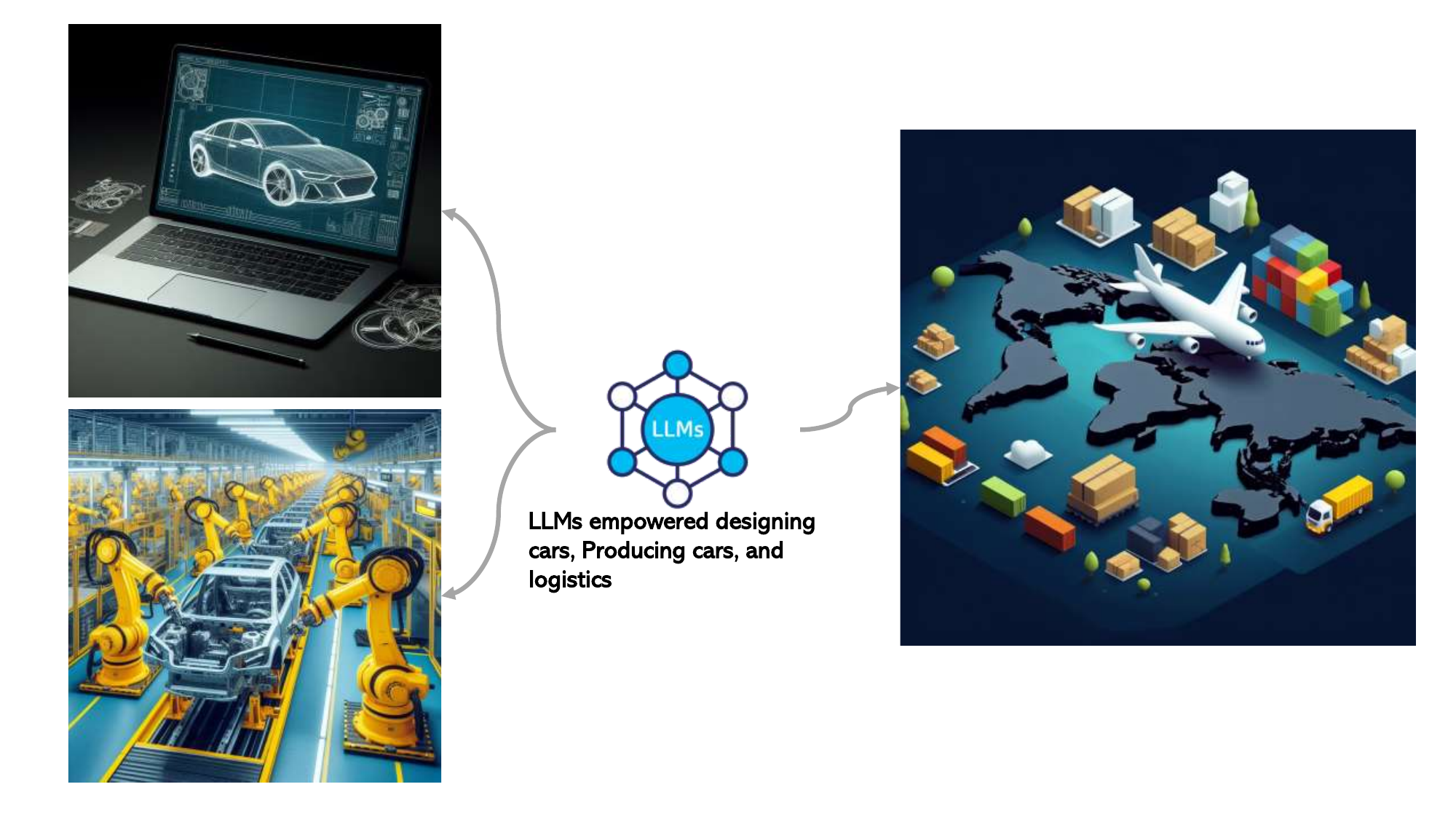}
\caption{Examples of LLMs being used to design cars, produce cars, and optimize logistics. Figures are generated by DALL·E 3.} 
\label{whole_process}
\end{figure*}

The integration of LLMs into the educational landscape within engineering and manufacturing education, has the potential to revolutionize both teaching and learning processes~\cite{pedro2019artificial}. For educators in these fields, LLMs can be powerful tools in enhancing educational goals by analyzing student performance data to set realistic and achievable targets. For students in engineering and manufacturing programs, LLMs provide personalized learning experiences by analyzing their learning patterns and preferences to offer customized content and activities. Figure \ref{exp_edu2} shows the interaction of LLMs for both educators and engineering students to improve education quality and experience.

In curriculum design, LLMs offer significant advantages by generating engaging and interactive course outlines, lesson plans, and assessment tasks specific to engineering and manufacturing subjects~\cite{gupta2023chatgpt,cooper2023examining}. Educators can access a vast array of technical content, enabling them to enrich their lessons beyond traditional textbooks and include up-to-date industry practices. Furthermore, LLMs assist in designing efficient assessment tools, providing personalized feedback, and offering real-time insights into student cognitive states through indicators like gaze and expression~\cite{zhai2023chatgpt}. This comprehensive support system enhances the overall teaching and learning process, making engineering education more inclusive and effective.

For engineering and manufacturing programs students, chatbots powered by LLMs ensure consistent communication, regular updates, and tailored suggestions related to technical course materials, which keeps students engaged and connected to the curriculum. Additionally, the vast array of resources made accessible through LLMs allows engineering students to broaden their knowledge and deepen their understanding of complex concepts such as mechanics, thermodynamics, and manufacturing processes.

The integration of LLMs into engineering education presents a promising avenue for enhancing both teaching and learning experiences. From personalized learning and instructional support to facilitating research and content creation, LLMs offer a range of benefits that can significantly contribute to the educational process. However, realizing their full potential requires careful consideration of implementation strategies and ethical considerations. As the technology continues to evolve, ongoing research and collaboration among educators, researchers, and AI developers will be key to leveraging LLMs effectively in the educational sector.

\subsection{Multimodality LLMs for Future Education}
Multimodality large language models (MLLMs) represent an advanced educational approach that integrates various forms of media and modes of learning to enhance the educational experience, including the fields of engineering and manufacturing. These models leverage the synergy of visual, auditory and textual learning modalities to create a rich, immersive, and interactive learning environment. By combining these different modes, MLLMs address diverse learning styles and needs, ensuring a more comprehensive understanding of complex engineering and manufacturing concepts.

With advancements in Visual Question Answering (VQA) and graphics generation, MLLMs have made remarkable strides in recent years. VQA was first introduced in \cite{antol2015vqa}, representing a novel approach at the intersection of computer vision and natural language processing. It focuses on creating systems capable of interpreting and responding to questions about visual content, such as images or videos. VQA systems combine image comprehension techniques, like object detection and scene interpretation, with natural language processing to parse and understand queries. By integrating multimodal features—visual and linguistic—they generate accurate, contextually relevant answers. Recently, numerous generative image models have gained popularity~\cite{creswell2018generative,he2021spatial,ramesh2021zero,van2016conditional,rombach2022high}. These models learn to capture the distribution of a dataset of images and can generate new images that align with the same distribution. Diffusion models, in particular, are trained by adding noise to an image and learning the denoising process that reconstructs the image from Gaussian noise. These technologies create a bridge between MLLMs and innovative teaching methods and application scenarios in future engineering education.

For visual learning in the manufacturing domain, video offers a means to capture the complexities of teaching and learning~\cite{lee2023multimodality}. Video analysis techniques have applications in a wide range of educational settings, from K-12 classrooms to higher education and professional development~\cite{gupta2023comprehensive}. For instance, \cite{van2020student} recorded classroom videos and, after processing, applied AI models to classify students into different emotional categories automatically. By analyzing students' emotions and moods, educators can adjust the difficulty of the coursework and engage with students who may be struggling, allowing for more personalized learning plans. MLLMs can also be utilized to enhance diagrams and schematics, offering clear and concise representations of complex systems and components, thereby helping students better understand the relationships and functions involved in engineering and manufacturing processes. 

Regarding flowcharts and process maps, visualizing workflows and processes allows students to comprehend the sequence of operations and the interdependencies among various stages of manufacturing. MLLMs can expedite the creation of animations and simulations, which educators use to illustrate dynamic processes such as machinery operations or material flow. These tools provide a real-time understanding of how systems function, bridging the gap between theoretical concepts and practical applications.

Audio analysis is widely supported by artificial intelligence
technology such as speech, voice, music and environmental sound recognition\cite{lee2023multimodality}. \cite{canovas2022analysis} developed an automated classroom audio analysis system that can distinguish the speakers’ identities (e.g., teacher, student) when there are multiple speakers simultaneously\cite{lee2023multimodality}. MLLMs analyze the requirement of students to gernerate the audio recordings of expert lectures on specific topics in engineering and manufacturing, which provides students with access to valuable insights and knowledge from industry professionals and academics, regardless of their location. Socratic Seminars can be established between students and MLLMs to encourage students to participate in structured, audio-based discussions where they can ask questions and explore complex topics deeply, fostering critical thinking and comprehension. Besides, Using multimedia presentations that combine audio explanations with visual aids, such as slides, diagrams, and videos, helps reinforce learning by engaging multiple senses simultaneously. All of these multimedia productions are potential application MLLMs can participate.

In conclusion, multimodality models represent a forward-thinking approach to engineering and manufacturing education, leveraging diverse learning modes to create an engaging, comprehensive, and practical educational experience. By adopting these models, educational institutions can better prepare students for the demands of the modern industrial landscape, equipping them with the knowledge and skills necessary for success in their careers.

\subsection{Challenges in Education}
While LLMs and MLLMs have demonstrated significant potential to enhance educational practices through personalized learning, automation of tasks, and improved access to knowledge, their integration into educational settings is not without challenges. These challenges extend beyond technical limitations, encompassing critical concerns related to instructional quality, student engagement, ethical considerations, and practical implementation. A nuanced understanding of these challenges is crucial for educators and policymakers as they seek to balance the opportunities offered by LLMs with the need to maintain the integrity and effectiveness of educational systems. The following sections will explore these key challenges in depth.

\begin{itemize}
    \item \textbf{Instruction Quality and Overreliance:} The integration of LLMs into educational settings holds the potential to significantly streamline instructional tasks, including the automation of question generation, essay grading, and the provision of instant feedback. However, the risk of overreliance on these models poses a critical challenge to maintaining high-quality education. The models, while capable of generating fluent text, may produce content that lacks the nuance or accuracy required in educational contexts, potentially introducing misinformation. For instance, LLMs such as GPT-3 and GPT-4 are prone to generating plausible-sounding but incorrect or misleading information, which could hinder the learning process rather than enhance it~\cite{yan2024practical}. Moreover, these models are subject to inherent biases derived from the datasets on which they are trained, which often reflect socio-cultural and racial biases, thus perpetuating educational inequalities. This raises concerns not only about the quality of education provided by LLMs but also about fairness and inclusivity in learning environments. Addressing these issues requires a careful balance between leveraging the efficiency of LLMs and ensuring that human oversight remains central to instruction.
    \item \textbf{Student Learning and Cognitive Engagement:} The ability of LLMs to provide immediate, seemingly accurate responses can have a profound effect on students’ learning trajectories, particularly by diminishing opportunities for deep cognitive engagement. While these models can serve as powerful tools for facilitating access to information, they may inadvertently encourage surface-level learning by offering ready-made solutions to complex problems. This reduces the necessity for students to engage in critical thinking or problem-solving, skills that are integral to the development of higher-order cognitive abilities. Research suggests that the widespread use of LLMs in education could lead to cognitive offloading, where learners become overly dependent on the model’s outputs, thereby weakening their ability to independently analyze and synthesize information~\cite{gan2023large}. In essence, while LLMs provide a valuable resource for supporting learning, their integration must be carefully managed to prevent a decline in active, self-directed learning, which is crucial for intellectual development.
    \item \textbf{Integrity and Bias:} The ethical implications of incorporating LLMs into educational environments are multifaceted, particularly in relation to issues of academic integrity, the propagation of biases, and concerns over privacy. The use of LLMs in the production of essays, reports, and other academic outputs raises concerns about plagiarism, as students may rely on these models to generate work that is not their own. This not only undermines the integrity of the educational process but also complicates the role of educators in assessing student performance~\cite{abd2023large}. Additionally, since LLMs are trained on vast datasets drawn from the internet, they tend to perpetuate societal biases, including racial, gender, and cultural biases. Such biases can be particularly problematic in educational contexts, where the promotion of equity and inclusivity is paramount. Furthermore, the issue of privacy arises from the extensive data collection required to train these models, raising concerns about how student data is collected, stored, and used. These ethical challenges necessitate the development of robust guidelines and regulatory frameworks to ensure that the use of LLMs in education is both responsible and equitable.
    \item \textbf{Transparency and Privacy:} The practical challenges associated with the deployment of LLMs in educational settings are significant and multifaceted. One major concern is the lack of transparency in how LLMs generate responses, which can make it difficult for educators and students alike to understand the basis of the information provided. This opacity undermines the ability to critically assess the reliability of the content produced by these models. Additionally, issues related to technological readiness further complicate the integration of LLMs in educational environments. Many existing LLM-based tools are not yet fully equipped to handle the diverse and complex needs of real-world educational settings, particularly in terms of scalability, replicability, and the customization required to address specific learning contexts~\cite{yan2024practical}. Privacy concerns also loom large, as the use of these models often involves processing sensitive data, including student assessments and personal information, which must be protected against breaches or misuse. Addressing these practical challenges requires advancements in model transparency, technological improvements, and the establishment of stringent data protection protocols.
\end{itemize}

\section{LLMs in Manufacturing Patent Management and Knowledge Management}

Patent management encompasses a multifaceted array of activities, including the search, application, maintenance, and commercialization of patents~\cite{grzegorczyk2020patent}. Technological support is indispensable for firms aiming to dissect and interpret market and technological trajectories~\cite{somaya2012patent}, pinpoint technology voids, and scout for potential collaborative partnerships. The deployment of valuation tools is crucial for estimating a patent's commercial viability and prospective market value~\cite{cao2013analysis}, facilitating a strategic approach in managing intellectual property~\cite{trappey2019patent} by enabling comprehensive analysis and informed decision-making regarding patent portfolios~\cite{ernst2016create}. The advent of large language models, characterized by their exceptional capabilities for processing and analyzing vast quantities of information, has facilitated the delegation of intricate and laborious text-related tasks to AI. This development underscores a significant shift in how data-intensive activities are managed, enabling more efficient handling of complex processes through automated systems.

LLMs are increasingly leveraged to automate the organization and analysis of extensive patent data. By extracting multi-level keywords from each patent document, LLMs enhance the entry process into patent management systems, significantly reducing the need for manual labor. Furthermore, these models conduct comprehensive analyses of multiple patents to ascertain their commercial potential and viability, providing strategic insights, suggesting potential business strategies, and research avenues, thereby enhancing operational efficiency.

Moreover, LLMs facilitate the patent application process, traditionally involving considerable repetitive and labor-intensive textual work~\cite{soranzo2017redesigning}. By generating preliminary application drafts based on provided keyword data and predetermined formats, LLMs streamline the drafting phase. Applicants can refine these drafts with minimal modifications, expediting the application process and substantially improving overall efficiency in patent filings.

The rise of LLMs marks a significant shift in information extraction and knowledge management in the manufacturing domain. Freire et al.~\cite{freire2024knowledge} explore the transformative potential of LLMs in redefining knowledge management within manufacturing. Their study introduces an innovative system that leverages LLMs to efficiently aggregate and disseminate knowledge from factory documents and expert insights. This approach streamlines information flow and enhances problem-solving capabilities on the production line. A key highlight is GPT-4V's superior performance over its counterparts, showcasing its exceptional capability in improving information retrieval in manufacturing environments. This comparison across various LLMs underscores the feasibility and transformative power of embedding such technologies into manufacturing settings, painting a future where knowledge extraction and management are seamlessly integrated into factory operations, fostering greater efficiency and informed decision-making. Recent advances in natural language processing enable more intelligent ways to support knowledge sharing in factories.

Freire et al.~\cite{kernan2024knowledge} demonstrate the efficacy of LLM-powered systems in streamlining knowledge extraction and sharing from extensive factory documents and expert insights. Through a meticulous user study in a factory environment, they reveal the system's adeptness in facilitating quicker access to information and enhancing problem-solving efficiency, though highlighting a general preference for human expertise. Their evaluation distinguishes GPT-4V as the most proficient model for such applications, paving the way for future explorations in LLM integration into industrial knowledge management.

Xia et al.~\cite{xia2024leveraging} take a novel approach by fine-tuning LLMs with error-assisted learning specific to manufacturing contexts, markedly improving the models' ability to understand and generate domain-relevant information. This technique addresses the critical challenge of adapting general-purpose LLMs to the specialized needs of the manufacturing sector, indicating a significant step towards more intelligent and responsive manufacturing systems.

Concurrently, Wang et al.~\cite{wang2023chatgpt} assess ChatGPT's utility in manufacturing, emphasizing its potential to generate coherent, structured responses supporting complex manufacturing tasks. Despite noting limitations in accuracy and reliability, their study highlights the necessity of critically evaluating ChatGPT's outputs to ensure their applicability and precision within industrial contexts.

Moreover, Meyer et al.~\cite{meyer2023llm} explore ChatGPT's integration into Knowledge Graph Engineering, showcasing its ability to automate and assist in creating, managing, and utilizing knowledge graphs. While acknowledging certain limitations, including the propensity for inaccuracies, they posit ChatGPT as a significant tool for reducing manual workload and enhancing the field's accessibility, contingent upon rigorous output validation.

These studies collectively illuminate the essential impact of LLMs on manufacturing sector's information extraction practices, signifying an evolving landscape where technology and domain-specific knowledge intersect to foster more efficient and intelligent manufacturing processes.

\textbf{}

While large language model can benefit patent and knowledge management in manufacturing by automating complex processes and enhancing decision-making, their integration into manufacturing operations underscores the critical need for trustworthy AI. Ensuring the reliability, safety, fairness, robustness, and explainability of these AI systems is paramount as they become more intertwined with essential manufacturing functions. By addressing these dimensions, manufacturers can fully leverage the transformative potential of LLMs and other AI technologies, optimizing processes and maintaining high standards of quality and safety.

\section{Trustworthy AI for Manufacturing}
The manufacturing industry is undergoing a significant transformation driven by the integration of artificial intelligence technologies, including large language models, computer vision, and robotics. As AI systems become more prevalent in various aspects of manufacturing, from design and production to quality control and supply chain management, ensuring their trustworthiness is crucial for their successful adoption and deployment.
Trustworthiness in AI encompasses multiple dimensions, such as reliability, safety, fairness, robustness, and explainability~\cite{liu2023trustworthy}.

In the context of manufacturing, these dimensions have specific implications:

\textbf{\textit{Reliability}} is essential for AI-powered quality control systems to consistently identify defects in products without false positives or false negatives. For example, computer vision algorithms used for inspecting printed circuit boards (PCBs) must accurately detect solder joint defects, component misalignments, or missing components~\cite{rajesh2024printed}. Similarly, LLMs used for analyzing customer feedback and market trends should provide accurate insights to inform product design and marketing strategies~\cite{loukili2023sentiment}.

\textbf{\textit{Safety}} is paramount in manufacturing environments where AI-controlled robots interact with human workers. Collaborative robots, or cobots, must operate safely, avoiding collisions and adapting to human movements~\cite{bragancca2019brief}. Advanced sensor fusion and real-time motion planning algorithms can enhance the safety of human-robot interaction~\cite{zhang2022human}. Additionally, AI systems must respect data privacy and security, protecting sensitive information such as product designs, process parameters, and employee records.

\textbf{\textit{Fairness}} is crucial in AI-assisted decision-making processes in manufacturing, such as production scheduling, resource allocation, and employee performance evaluation. Scheduling algorithms should ensure equitable allocation of machine time and resources across different products and orders, considering factors such as deadlines, priorities, and setup times~\cite{amer2022optimized}. Performance evaluation systems should base their recommendations on objective metrics and avoid biases related to gender, race, or other protected attributes~\cite{pagano2023bias}.

\textbf{\textit{Robustness}} is critical for AI systems in manufacturing to maintain performance under varying conditions. For instance, LLMs used for generating product designs or process instructions should be resilient to variations in input requirements or user preferences. Anomaly detection models for predictive maintenance should adapt to changes in operating conditions, such as variations in raw material properties or environmental factors. Robust optimization techniques can help AI models generate production plans that are less sensitive to uncertainties in demand, supply, or processing times.

\textbf{\textit{Explainability}} is important for building trust and accountability in AI-driven manufacturing processes. Plant managers and operators should be able to understand the reasons behind AI-generated production schedules, maintenance recommendations, or quality control decisions. Techniques such as feature importance analysis, rule extraction, or counterfactual explanations can provide insights into the factors influencing AI outputs. Explainable AI can also facilitate compliance with regulations and audits by providing transparent and interpretable records of AI-assisted decisions, such as those made by LLMs in supply chain management.

To operationalize trustworthy AI in manufacturing, practitioners can leverage frameworks and methodologies from the broader AI ethics and responsible AI literature, adapting them to the specific needs of manufacturing. This includes establishing governance structures, conducting impact assessments, and engaging in multi-stakeholder dialogues to align AI systems with industry standards, best practices, and societal expectations.

Collaborative efforts among industry, academia, and policymakers will be essential to develop and adopt best practices for trustworthy AI in manufacturing. By proactively addressing reliability, safety, fairness, robustness, and explainability, manufacturers can harness the power of AI to optimize processes, enhance product quality, improve worker safety and satisfaction, and achieve sustainable competitive advantages in the rapidly evolving manufacturing landscape.

\section{Discussions and Conclusion}

\subsection{Limitations}
\begin{itemize}
\item \textbf{Domain-Specific Knowledge Acquisition:} Deploying LLMs within the manufacturing sector poses the challenge of acquiring domain-specific knowledge. Manufacturing involves specialized terminologies, regulatory frameworks, and dynamic market conditions. Ensuring LLMs can effectively comprehend and process this intricate data is crucial for their successful application.
\item \textbf{High Expectations for Engineering Outcomes:} The proximity of manufacturing to engineering disciplines sets high expectations for LLM-generated outcomes. Manufacturing processes require precise calculations, adherence to strict tolerances, and compliance with safety regulations. While LLMs have shown significant potential for logical reasoning, consistently meeting these stringent requirements remains challenging.
\item \textbf{Efficient Integration into Downstream Tasks:} Although LLMs have shown promise in various manufacturing applications, integrating them efficiently into downstream tasks is challenging. Manufacturing workflows involve complex interactions between multiple systems and stakeholders. Seamlessly incorporating LLMs requires careful consideration of data flows, user interfaces, and system architectures.
\end{itemize}

\subsection{Future Directions}
\begin{itemize}
\item \textbf{Enhancing Domain Adaptation Techniques:} To address the challenge of domain-specific knowledge acquisition, researchers are exploring advanced domain adaptation techniques for LLMs. This involves training models on curated corpora of manufacturing-related data, including technical documents, industry standards, and operational manuals. Exposing LLMs to a wide range of relevant information can significantly improve their ability to understand and generate content specific to the manufacturing domain.
\item \textbf{Developing Hybrid Approaches:} While LLMs excel at processing unstructured data and generating human-like text, they may struggle with tasks requiring precise numerical computations or strict logical reasoning. Researchers are investigating hybrid approaches combining LLMs with traditional rule-based systems or numerical optimization techniques. Leveraging the strengths of each approach can develop more robust and reliable solutions for manufacturing applications.
\item \textbf{Collaborative Development and Evaluation Frameworks:} Efficient integration of LLMs into manufacturing workflows requires collaborative development and evaluation frameworks. These frameworks should provide standardized interfaces for data exchange, model deployment, and performance monitoring. Establishing common protocols and best practices can accelerate the adoption of LLMs in manufacturing and more effectively measure and optimize their impact.
\item \textbf{Explainable AI for Manufacturing:} As LLMs are increasingly applied to critical decision-making processes in manufacturing, the need for explainable AI becomes paramount. Researchers are exploring techniques to enhance the interpretability of LLMs, allowing users to understand the reasoning behind the model's outputs. This includes developing methods for generating human-readable explanations, visualizing attention mechanisms, and quantifying uncertainty in model predictions. Improving the transparency and accountability of LLMs in manufacturing can strengthen trust in these systems and encourage their adoption.
\end{itemize}

As of the completion of this paper, OpenAI has released its latest large model, o1. The o1 model has demonstrated outstanding performance in complex reasoning tasks, including coding, scientific problem-solving, and technical language processing. With an 83.3\% success rate in solving complex programming challenges and strong capabilities in generating technical reports, o1 can enhance manufacturing processes by improving code precision, automation, and data analysis. Integrating o1 into manufacturing pipelines can optimize production, supply chain management, and design processes. Additionally, o1's proficiency in multi-modal integration enables real-time analysis of textual, visual, and technical data, facilitating efficient communication and decision-making across departments.

\subsection{Conclusion}
This study has highlighted the immense potential of LLMs in revolutionizing various aspects of the manufacturing sector. Through a comprehensive exploration of LLMs' applications in domains such as engineering design, supply chain management, quality control, cost control, and robotics, we have demonstrated their ability to drive innovation, efficiency, and optimization in manufacturing processes.

The exceptional performance of LLMs, particularly GPT-4V, in tasks related to data analysis, code generation, and zero-shot learning underscores their transformative potential. However, the successful deployment of LLMs in manufacturing also faces challenges, including the need for domain-specific knowledge acquisition and the high expectations for precision and reliability in engineering outcomes.

As the manufacturing industry continues to evolve, the integration of LLMs is expected to play a crucial role in shaping its future. Ongoing research efforts in areas such as domain adaptation, hybrid approaches, and explainable AI will be instrumental in unlocking the full potential of LLMs in manufacturing.

Looking ahead, the symbiotic relationship between LLMs and the manufacturing sector is poised to drive significant advancements. By harnessing the power of these AI technologies, manufacturers can gain valuable insights, streamline operations, and enhance decision-making processes. The future of manufacturing lies in the successful integration of LLMs, and this study serves as a stepping stone towards realizing that vision.

\section*{Acknowledgment}

The authors would like to express their sincere gratitude to Dr. S. Jack Hu for his insightful guidance and constructive feedback throughout the development of this work. His expertise and suggestions were invaluable in refining the ideas presented in this paper.



\bibliographystyle{alpha}
\bibliography{sample}

\end{document}